\documentclass[11pt,letterpaper]{article}
\usepackage[T1]{fontenc}
\usepackage[utf8]{inputenc}
\usepackage{lmodern}
\usepackage[margin=1in]{geometry}
\usepackage{amsmath,amssymb}
\usepackage{graphicx}
\usepackage{booktabs,longtable,array,calc}
\usepackage{microtype}
\usepackage{xurl}
\usepackage[hidelinks]{hyperref}
\usepackage{caption}
\hypersetup{pdftitle={From Isolated Tasks to Structured Capabilities: A Multilayer Taxonomy for Large Language Models},pdfauthor={Shixin Fang; Jiachen Wo; Sihang Jiang; Wenjuan Qin; Yanghua Xiao}}

\title{From Isolated Tasks to Structured Capabilities: A Multilayer Taxonomy for Large Language Models}
\author{Shixin Fang, Jiachen Wo, Sihang Jiang\textsuperscript{*},
Wenjuan Qin\textsuperscript{*}, and Yanghua Xiao\\[4pt]
\small Fudan University\\[3pt]
\footnotesize \texttt{fangshixin@fudan.edu.cn}; \texttt{25213050400@m.fudan.edu.cn}\\
\footnotesize \texttt{jiangsihang@fudan.edu.cn}; \texttt{qin\_wenjuan@fudan.edu.cn};
\texttt{shawyh@fudan.edu.cn}\\[3pt]
\footnotesize \textsuperscript{*}Co-corresponding authors}
\date{}

\begin{document}
\maketitle

\begin{abstract}
Large language model (LLM) evaluation spans diverse tasks and benchmarks, yet evidence remains organized around tasks rather than the capabilities they probe. This fragmentation limits cross-study comparison, obscures capabilities tasks recruit, and makes coverage gaps difficult to identify.

We introduce a multi-layer taxonomy of 14 capability domains and 91 subskills across Primitive, Constructed, and Integrative layers. Human cognitive science guides capability definition and organization, not LLM architecture. Layer assignments draw on developmental precedence and hypothesized functional support, while human-origin constructs are adapted to observable model behavior.

To demonstrate operational utility, we screened 31,505 papers from ACL, AAAI, ICML, and NeurIPS between 2023 and 2025 and mapped 15,934 LLM-focused papers through multi-model annotation, consensus, and arbitration. Direct research attention concentrated on \emph{Language-Semantic Competence} (3,551; 22.3\%), \emph{Reasoning} (3,388; 21.3\%), \emph{Planning and Decision-Making} (2,149; 13.5\%), and \emph{Perception} (1,954; 12.3\%), whereas six domains appeared in fewer than 2\% of papers. Within domains, the most frequent subskill had a median prevalence of 97.9\% and appeared in at least 90\% of papers in 10 of 14 domains. \emph{Language-Semantic Competence} and \emph{Reasoning} formed the highest-volume pair (n = 1,864; 11.7\%; lift = 2.47), whereas \emph{Theory of Mind} and \emph{Social Reasoning and Interaction} showed the highest lift among pairs with at least 20 co-occurrences (n = 62; lift = 30.84).

By shifting the unit of analysis from isolated tasks to structured capabilities, the taxonomy supports research organization, coverage audits, evaluation interpretation, and testable hypotheses for diagnosis, training, and transfer.
\end{abstract}

\section{Introduction}

Human cognition is often described as a system of interacting foundational and higher-order processes that develop over time (Diamond, 2013). Contemporary large language models (LLMs), however, often show a very different capability profile.

State-of-the-art LLMs now achieve strong performance on difficult knowledge and multi-step reasoning evaluations (OpenAI, 2025; Wei et al., 2022). At the same time, they can remain brittle on elementary rule-governed operations, including counting and simple symbolic transformations (McCoy et al., 2024). This fragility can persist even after apparent success: performance may degrade when task premises or rules are counterfactually modified (Wu et al., 2024). We refer to this pattern ability inversion: a model may look strong on tasks usually regarded as cognitively advanced while remaining unreliable on basic operations that humans often perform consistently once the relevant rules are understood. For LLM evaluation, ability inversion turns an apparent anomaly into a measurement problem, because benchmark success alone does not show whether the underlying capability is stable, transferable, or compositionally supported.

This measurement problem reflects a structural limitation in how LLM capabilities are represented, studied, and evaluated. Most benchmarks are task-centric: they measure performance on particular datasets, tasks, or scenarios without specifying which capability is being probed, which additional capabilities the task recruits, or what a failure implies about the model's broader capability profile. As a result, benchmark scores are often descriptive but not diagnostic. If a model fails a reasoning task, for example, the failure may reflect reasoning itself, limitations in memory or attention, or a breakdown in coordinating these capabilities. Existing evaluations more often tell us what a model scored than why it succeeded or failed.

A related consequence of task-centric evaluation is the uneven distribution of research attention across cognitive capabilities. A systematic review of 445 LLM benchmarks from leading NLP and ML conferences finds that reasoning alone accounts for 18.5\% of all evaluated phenomena, with coverage concentrated in a handful of domains (Bean et al., 2025). Using Bloom's Taxonomy, a complementary analysis finds that existing datasets cluster around lower-level remembering and understanding while metacognitive and creative reasoning stay underrepresented (Zhang et al., 2025). Psychological and human-centered constructs remain fragmented and less standardized in current LLM evaluation (Chang et al., 2024; Ye et al., 2025). This raises the empirical question of how research attention is distributed across the broader cognitive capability space, including for example, social, cultural, or moral domains. Such concentration may also shape subsequent evaluation and development priorities, leaving less frequently studied capabilities comparatively undermeasured and difficult to diagnose.

Existing approaches partially address these diagnostic and coverage problems. Recent benchmark frameworks have expanded the breadth and realism of LLM evaluation, including expert-level knowledge assessment, coding, multimodal reasoning, and agent-oriented evaluation, as in Humanity's Last Exam (Phan et al., 2025), LiveCodeBench (Jain et al., 2025), and BrowseComp (Wei et al., 2025). However, many remain organized around tasks or deployment scenarios rather than around a structured account of the capabilities those evaluations target. Data-driven approaches can reveal empirical similarity or transfer relations among tasks and skills (Chen et al., 2023), but these relations do not by themselves provide a broad cognitive organization of the capability space. Cognitive-inspired evaluations such as CogBench (Coda-Forno et al., 2024), CogLM (Wang et al., 2024), CognitivEval (de Langis et al., 2025), and NeuroCognition (Haznitrama et al., 2026) show the value of psychologically grounded constructs for interpreting LLM behavior beyond aggregate scores. Their primary unit of analysis, however, remains the measured construct or task rather than systematic relations across capability domains. More recent frameworks, including CDT (Mo et al., 2025) and the cognitive framework of Burnell et al.~(2026), organize capabilities at a more structured level. Burnell et al.~(2026) explicitly distinguish basic from composite faculties and acknowledge that cognitive faculties interact and build on one another, but neither their framework nor CDT jointly uses developmental precedence and hypothesized functional support to define a multi-layer taxonomy. Holistic frameworks such as HELM broaden evaluation beyond accuracy by jointly considering scenarios and multiple desiderata, but their organizing units remain scenarios and metrics rather than cognitive capability relations (Liang et al., 2023). Taken together, these advances leave a central representational problem unresolved: how to organize fragmented task-level evidence into a common capability space that supports systematic comparison, coverage analysis, and diagnostic hypothesis generation.

Human cognition provides a principled reference for addressing this gap. LLMs may fail on tasks that humans find elementary while succeeding on tasks that are difficult for most people. The asymmetry suggests that human cognition may generate useful structural hypotheses about capability gaps of LLMs. Converging developmental, neurocognitive, and cognitive traditions portray cognition as involving interacting processes with heterogeneous developmental trajectories and coordinated functional support (Diamond, 2013; Gogtay et al., 2004; Miller \& Cohen, 2001; Piaget, 1952; Vygotsky, 1978). Human cognitive research therefore offers three resources for taxonomy construction: a broad candidate capability space, evidence for organizing relationships among capabilities, and established constructs that can be adapted into observable targets for LLM research.

These resources lead to three design decisions for studying LLMs: which capabilities to include, how to organize them, and how to make them operational for LLM research. They motivate three principles. First, human cognitive science provides a broad reference space for identifying capabilities beyond existing task and benchmark categories. Second, capabilities should be organized according to developmental precedence and hypothesized functional support rather than represented as a flat list of independent skills. Third, human cognitive constructs must be adapted into observable and operationally useful targets for studying and evaluating LLMs. These principles do not imply that LLMs reproduce human cognitive architectures or follow human developmental trajectories. Instead, they define a structured capability space that can organize existing research, reveal patterns of coverage and neglect, and generate testable hypotheses about relationships among capabilities.

Building on these principles, we propose a multilayer taxonomy that represents LLM cognitive capabilities as a structured space comprising 14 domains across three layers: Primitive (L1), Constructed (L2), and Integrative (L3) Cognitive Capabilities (Figure 1). The taxonomy represents candidate vertical relations across layers and lateral relations within layers. Rather than organizing LLM research solely by application domains, tasks, benchmarks, or output similarity, it allows research artifacts to be mapped according to the capabilities they explicitly study and the subconstructs through which those capabilities are operationalized. This structured representation supports research and benchmark mapping, reveals gaps and imbalances in capability coverage, and generates testable hypotheses for targeted follow-up evaluation.

This paper makes three contributions. First, it introduces a multi-layer taxonomy that organizes 14 cognitive capability domains and 91 subskills into Primitive, Constructed, and Integrative layers, with layer assignments informed by developmental precedence and hypothesized functional support in human cognition. Second, it operationalizes the taxonomy as an annotation codebook and scalable mapping procedure that distinguishes capabilities studied as primary targets from those appearing only indirectly or contextually. Third, it maps 15,934 LLM-focused papers, revealing systematic asymmetries in which capability domains receive attention, which subskills dominate within them, and which capabilities are studied together. These contributions provide a common basis for comparing evaluation evidence, identifying blind spots in the capability landscape, and generating testable diagnostic hypotheses.

% Alt text: Three-layer taxonomy of 14 cognitive capability domains, organized into Primitive, Constructed, and Integrative layers with candidate support relations.
\begin{figure}[p]
\centering
\includegraphics[width=0.94\linewidth,height=0.82\textheight,keepaspectratio]{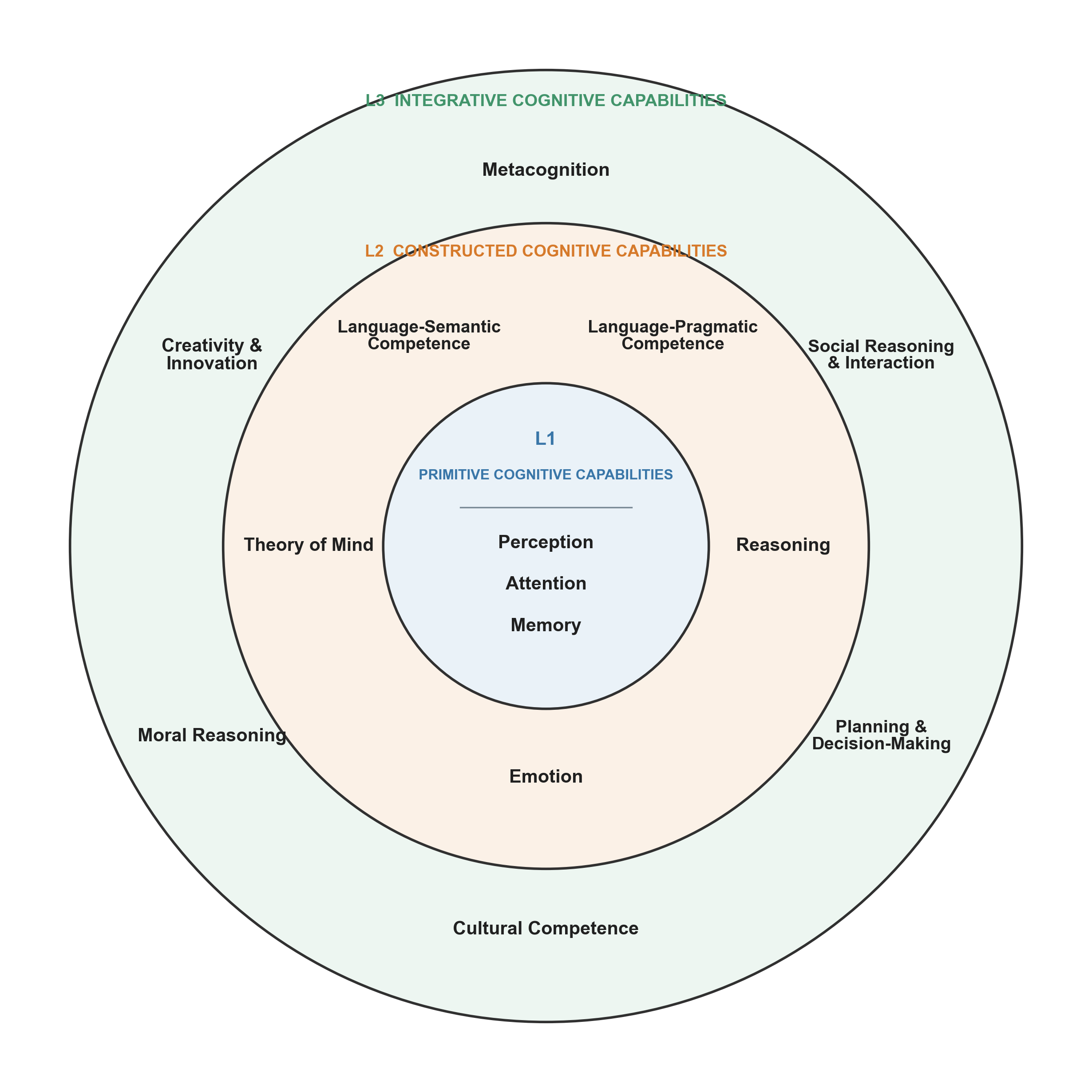}
\caption{The multi-layer taxonomy of cognitive capabilities.}
\label{fig:1}
\end{figure}

\section{Taxonomy Design Principles and Construction}

This section describes the design principles and construction process of the proposed taxonomy. It is organized around three questions: What capability space should serve as a reference for LLM research? How should capabilities within that space be organized in relation to one another? And how should constructs derived from human cognition be reformulated as observable and operationally useful targets for LLM research? These questions motivate three design principles: human cognition as a reference space for general cognitive capability, multi-layer organization, and LLM-specific adaptation and operationalization.

\subsection{Scope and Design Objective}

The taxonomy is a structured capability space for describing, organizing, and relating cognitive capabilities in LLMs. It is not a model of LLM internal architecture and not a benchmark suite. Instead, it provides a representational layer that can support literature synthesis, evaluation design, capability diagnosis, and development-oriented analysis without being defined by any single application for LLMs.

Research on human cognition provides three resources for our taxonomy construction: a well-developed vocabulary for distinguishing capabilities, empirical and theoretical accounts of their functional relationships, and established measurement traditions that can inform construct operationalization and task design. The taxonomy therefore represents observable capability constructs and candidate functional relationships among them.

\subsection{Design Principles}

The three principles correspond to three distinct design decisions: how to define the initial capability space, how to organize capabilities within that space, and how to translate human cognitive constructs into an LLM-compatible conceptual and operational taxonomy.

\subsubsection{Principle 1: Human Cognition as a Reference Space for General Cognitive Capability}

Human cognitive science provides a mature and broad reference space for identifying the capabilities relevant to general cognitive competence. We therefore derived the initial candidate space from constructs that recur across cognitive psychology, cognitive science, cognitive neuroscience, and adjacent fields.

Our starting sources included widely used integrative textbooks and handbooks, such as \emph{Cognitive Psychology: A Student's Handbook} (Eysenck \& Keane, 2020), \emph{Cognition: Exploring the Science of the Mind} (Reisberg, 2013), and \emph{Cognitive Neuroscience} (Banich \& Compton, 2023). Across these sources, a relatively stable set of constructs recurs, including perception, attention, memory, language, reasoning and problem solving, judgment and decision-making, and cognitive control.

This reference space covers capabilities that become increasingly relevant as LLMs are deployed in agentic, educational, advisory, interpersonal, and socially consequential settings. A taxonomy intended to characterize broadly capable systems must therefore extend beyond language proficiency, factual knowledge, and abstract reasoning. Broader literatures additionally foreground emotion, creativity, metacognition, planning, social cognition, and the cultural and developmental contexts of cognition.

\subsubsection{Principle 2: Multi-Layer Organization}

The second principle concerns how capabilities are organized in relation to one another. Layer assignment draws on two complementary forms of evidence: human developmental precedence and functional support. Developmental precedence concerns whether a capability emerges relatively early or develops through the coordination of capacities established earlier, whereas functional support concerns whether one capability provides representations or operations on which another capability builds. These criteria are used as convergent organizational evidence rather than as claims that LLMs reproduce human developmental trajectories or cognitive architectures.

Developmental precedence provides evidence for distinguishing relatively foundational capabilities from those involving greater coordination and integration. Broad theories of cognitive development characterize development as the progressive construction of increasingly complex skills through the coordination of skills established at preceding levels (Fischer, 1980). For example, early manifestations of attention and memory are measurable in infancy through visual-attention and recognition-memory paradigms (Reynolds \& Richards, 2019), whereas the planning component of planning and decision-making follows a more protracted trajectory, with performance on difficult strategic-planning problems continuing to improve through late adolescence and into early adulthood (Albert \& Steinberg, 2011). Contemporary accounts nevertheless emphasize that developmental trajectories are overlapping, domain-sensitive, and nonlinear rather than constituting a single invariant sequence of capability acquisition (Gauvain, 2022; Houdé \& Borst, 2022). Developmental precedence is therefore treated as convergent evidence for layer assignment, not as a claim that all capabilities or LLMs develop through fixed human stages.

Functional support provides a complementary criterion for layer assignment. Capability A is treated as supporting B when performance on B typically recruits representations or operations associated with A. Task co-occurrence alone is insufficient. For example, \emph{Language-Semantic Competence} builds on \emph{Memory} when discourse-level meaning requires information introduced earlier to be maintained and integrated with subsequent linguistic input (Baddeley, 2003; Just \& Carpenter, 1992). \emph{Moral Reasoning} similarly builds on \emph{Theory of Mind} when evaluating an action requires integrating an agent's beliefs and intentions rather than judging outcomes alone (Fadda et al., 2016; Moran et al., 2011). More generally, research on executive and hierarchical control supports the distinction between relatively basic operations and more complex capacities that coordinate them (Badre \& Nee, 2018; Diamond, 2013). These relations do not imply that A is sufficient for B, that every manifestation of B depends equally on A, or that LLMs implement the corresponding human mechanisms.

The resulting layers therefore reflect both developmental emergence and functional organization. Primitive Cognitive Capabilities provide relatively early-emerging and broadly reusable representations and operations. Constructed Cognitive Capabilities organize these foundations into coherent linguistic, inferential, affective, and mental-state capacities. Integrative Cognitive Capabilities coordinate multiple primitive and constructed capabilities across contexts, goals, and extended courses of action.

\subsubsection{Principle 3: LLM-Specific Adaptation and Operationalization}

The third principle concerns how constructs drawn from human cognition are adapted for LLM research. Candidate constructs were filtered, reformulated, consolidated, or extended according to their relevance to observable model behavior, capability analysis, and deployment contexts.

Constructs without meaningful behavioral analogues were excluded. Fine motor control, for example, is central to human functioning but falls outside the scope of this taxonomy. Constructs referring to subjective experience, motivation, or internal states are retained only insofar as they can be expressed through observable behavior, explicit model outputs, or system-level functional analogues; their inclusion does not imply that LLMs possess the corresponding human experiences or mechanisms. Accordingly, \emph{Perception} refers to observable input-processing functions rather than biological sensory transduction or subjective experience. Related human constructs were consolidated when they served overlapping functional or analytic roles. Language was divided into \emph{Language-Semantic Competence} and \emph{Language-Pragmatic Competence} because they capture distinct dimensions of model capability: the former concerns lexical, compositional, referential, and discourse-level meaning, whereas the latter concerns context-dependent, implied, nonliteral, audience-sensitive, and socially situated language use. Language evaluation should therefore distinguish distributional competence from the ability to use meaning in physical and social context (Bisk et al., 2020).

LLM-specific adaptation also motivated the inclusion of capabilities that are not always treated as independent domains in cognition but are increasingly important in socially consequential deployment. For example, \emph{Cultural Competence} is retained because LLMs operate across diverse norms, practices, values, and communicative contexts. \emph{Moral Reasoning} is retained because model outputs are increasingly used in safety-relevant and norm-sensitive settings.

Finally, each domain is defined at a level that can be linked to observable model behavior, tasks, benchmarks, and research artifacts. In this paper, a domain is operationalizable if annotators and researchers can identify it, distinguish it from adjacent domains, and apply it consistently in literature mapping, benchmark analysis, capability diagnosis, and model-development research.

\subsection{Construction Process}

We developed the taxonomy through four iterative phases.

\textbf{Phase 1: Candidate identification.} We identified an initial pool of capabilities from recurring constructs in major textbooks and handbooks across cognitive psychology, cognitive science, cognitive neuroscience, and related fields.

\textbf{Phase 2: Construct elaboration.} Candidate constructs were provisionally grouped to guide domain-specific searches. Ten research assistants searched Web of Science for seminal and widely cited theoretical and assessment literature published over the preceding 50 years. The resulting evidence was synthesized to refine construct definitions, domain boundaries, and candidate subskills.

\textbf{Phase 3: Domain finalization and hierarchical organization.} Based on this synthesis, constructs were retained, consolidated, or excluded according to conceptual distinctiveness, relevance to LLM research and deployment, and functional role. The resulting taxonomy comprised 91 canonical subskills across 14 domains. Domains were assigned to three layers using developmental precedence and functional support, with greater weight given to functional support when the two criteria diverged.

\textbf{Phase 4: LLM-specific operationalization.} Domain and subskill definitions were adapted to observable model behavior, explicit outputs, and system-level functional analogues. This phase produced the final taxonomy and annotation codebook used in the literature-mapping analysis, including inclusion criteria, evidence-strength rules, and negative mapping rules for engineering terms such as attention, memory, cache, routing, and reward shaping.

Although presented sequentially, the process was iterative, with later synthesis and operationalization informing revisions to domain boundaries, subskill definitions, and layer assignments. The process establishes the taxonomy's conceptual and operational basis but does not independently validate its proposed dependency relations.

\section{A Multi-layer Taxonomy of Cognitive Capabilities}

The proposed taxonomy comprises 14 cognitive capability domains organized into three layers (Figure 1). The Primitive layer includes \emph{Perception, Attention, and Memory}. The Constructed layer includes \emph{Language-Semantic Competence, Language-Pragmatic Competence, Reasoning, Emotion,} and \emph{Theory of Mind.} The Integrative layer includes \emph{Metacognition, Social Reasoning and Interaction, Planning and Decision-Making, Creativity and Innovation, Cultural Competence,} and \emph{Moral Reasoning.} The domains are units of analysis for evaluation and literature mapping; they are not intended to form a mutually exclusive partition of cognition.

\subsection{The Three Capability Layers}

The \emph{Primitive layer} comprises \emph{Perception}, \emph{Attention}, and \emph{Memory}. These domains capture the foundational functions of constructing representations from input, selecting task-relevant information, and retaining and retrieving information across contexts and timescales. The term \emph{Primitive} denotes their foundational role in the taxonomy, not conceptual simplicity or full maturity at birth. For LLM-centered systems, \emph{Perception} covers the functional encoding, extraction, integration, organization, recognition, and adaptive interpretation of linguistic, symbolic, sensory, or multimodal input, rather than subjective sensory experience.

The \emph{Constructed layer} comprises \emph{Language-Semantic Competence}, \emph{Language-Pragmatic Competence}, \emph{Reasoning}, \emph{Emotion}, and \emph{Theory of Mind}. The two language domains distinguish the processing of lexical, compositional, referential, and discourse-level meaning from context-dependent and socially situated language use. These capabilities organize perceptual, attentional, mnemonic, and representational resources into coherent linguistic, inferential, affective, and mental-state processing.

The \emph{Integrative layer} comprises \emph{Metacognition}, \emph{Social Reasoning and Interaction}, \emph{Planning and Decision-Making}, \emph{Creativity and Innovation}, \emph{Cultural Competence}, and \emph{Moral Reasoning}. These capabilities typically require the coordinated recruitment of multiple Primitive and Constructed capabilities and, in some cases, other Integrative capabilities. They are particularly relevant to complex, agentic, interpersonal, culturally situated, and norm-sensitive applications of LLMs.

\subsection{Capability Definitions}

Table 1 provides concise, LLM-oriented definitions of the 14 domains. Each definition specifies the domain's primary evaluative target rather than exhaustively covering every construct associated with it. The appendix provides the complete 14-domain, 91-subskill construct-label scheme used in the analysis, together with the principal annotation and cross-domain interpretation rules.

\begingroup
\small
\begin{longtable}[]{@{}
>{\raggedright\arraybackslash}p{(\linewidth-4\tabcolsep)*\real{0.160}}
>{\raggedright\arraybackslash}p{(\linewidth-4\tabcolsep)*\real{0.240}}
>{\raggedright\arraybackslash}p{(\linewidth-4\tabcolsep)*\real{0.600}}@{}}
\caption{Overview of the 14 Cognitive Capability Domains}\tabularnewline
\toprule
Layer & Capability & Concise LLM-oriented definition \\
\midrule
\endfirsthead
\toprule
Layer & Capability & Concise LLM-oriented definition \\
\midrule
\endhead
Primitive & Perception & The functional encoding, extraction, integration, organization, recognition, and adaptive interpretation of structure in linguistic, symbolic, or multimodal input. \\
& Attention & The orienting, selection, and sustained prioritization of task-relevant information while resisting irrelevant, distracting, or competing input. \\
& Memory & The encoding, temporary maintenance and manipulation, retention, retrieval, and prospective use of information, experiences, skills, and intended actions across timescales. \\
Constructed & Language-Semantic Competence & The interpretation and generation of lexical meaning, grammatical and compositional structure, reference, logical-semantic relations, and discourse-level coherence. \\
& Language-Pragmatic Competence & The interpretation and generation of context-dependent meaning, including implicature, presupposition, indirect speech acts, figurative language, politeness, stance, contextual appropriateness, and sociocultural meaning. \\
& Reasoning & The derivation and evaluation of deductive, inductive, probabilistic, causal, analogical, and relational inferences. \\
& Emotion & The observable recognition, interpretation, response to, regulation, acceptance-oriented handling, and use of emotional information concerning oneself, others, or situations. \\
& Theory of Mind & The representation and inference of others' emotions, desires, intentions, perceptual perspectives, knowledge states, and beliefs. \\
Integrative & Metacognition & The observable knowledge, monitoring, evaluation, experience, beliefs, and regulation of cognition at individual, collaborative, and human--AI levels. \\
& Social Reasoning and Interaction & The interpretation and coordination of social information across cues, persons, mental states, communicative intentions, roles, rules, norms, relationships, and collaborative interactions. \\
& Planning and Decision-Making & An iterative, goal-directed process involving problem framing, self-appraisal, evidence acquisition and evaluation, option generation and selection, action sequencing, implementation, stakeholder and risk management, outcome monitoring, and adaptive revision. \\
& Creativity and Innovation & The exploratory generation, flexible development, proactive implementation, and iterative refinement of ideas or solutions that are novel, useful, and contextually appropriate. \\
& Cultural Competence & Critical reflection on cultural assumptions and power, integration of contextual cultural knowledge, adaptive communication, culturally responsive practice, and power-conscious engagement. \\
& Moral Reasoning & The evaluation of moral situations and the translation of moral judgment into consistent commitment and action, including consideration of principles, norms, consequences, responsibilities, and contextual pressures. \\
\bottomrule
\end{longtable}
\endgroup

Several general principles guide the interpretation of these definitions. All domains refer to observable behavioral or functional manifestations rather than claims about subjective experience or internal cognitive mechanisms. The domains are analytically distinguishable but not strictly mutually exclusive, because complex cognitive performance often integrates representations and processes associated with multiple capabilities.

\section{Illustrative Mapping of Recent LLM Capability Research}

This section applies the 14-domain taxonomy to recent LLM research and addresses three questions: How are cognitive capability domains distributed across recent LLM research (RQ1)? How are these domains represented through their fine-grained subskills (RQ2)? Which capability domains are studied together (RQ3)? The analysis is intended as an illustrative, reproducible use case for the taxonomy rather than as a complete census of all LLM capability research.

\subsection{Methods: Taxonomy-Based Literature Mapping}

\subsubsection{The Corpus}

We screened 31,505 papers published at ACL, AAAI, ICML, and NeurIPS between 2023 and 2025 (Figure 2). A paper was retained if an LLM or LLM-based agent was a central object of study and the paper substantively examined, evaluated, explained, or improved at least one capability represented in the taxonomy. Multimodal and agentic studies were included when an LLM was central to the system or the paper's main claims. Studies focused exclusively on non-LLM diffusion models, reinforcement-learning agents, or embodied systems were excluded. This procedure yielded an analytic corpus of 15,934 LLM-focused papers. Table 2 summarizes the screening pool and retained analytic corpus.

\begingroup
\footnotesize
\begin{longtable}[]{@{}
>{\raggedright\arraybackslash}p{(\linewidth-6\tabcolsep)*\real{0.140}}
>{\raggedright\arraybackslash}p{(\linewidth-6\tabcolsep)*\real{0.220}}
>{\raggedright\arraybackslash}p{(\linewidth-6\tabcolsep)*\real{0.190}}
>{\raggedright\arraybackslash}p{(\linewidth-6\tabcolsep)*\real{0.450}}@{}}
\caption{Screening Pool and LLM-Focused Analytic Corpus}\tabularnewline
\toprule
\textbf{Venue} & \textbf{Screened papers} & \textbf{LLM-focused analytic papers} & \textbf{Retained share} \\
\midrule
\endfirsthead
\toprule
\textbf{Venue} & \textbf{Screened papers} & \textbf{LLM-focused analytic papers} & \textbf{Retained share} \\
\midrule
\endhead
AAAI & 6,937 & 3,347 & 48.2\% \\
ACL & 4,262 & 2,920 & 68.5\% \\
ICML & 7,768 & 3,266 & 42.0\% \\
NeurIPS & 12,538 & 6,401 & 51.1\% \\
\textbf{Total} & \textbf{31,505} & \textbf{15,934} & \textbf{50.6\%} \\
\bottomrule
\end{longtable}
\endgroup

\subsubsection{Annotation Agent and Label Schema}

We build a taxonomy annotation agent that assigns cognitive capability labels to LLM research papers. For each paper the agent emits two categories of labels: (i) capability labels drawn from the 14 conceptual domains organized across the three layers of the taxonomy (Primitive, Constructed, Integrative), and (ii) subskill labels drawn from the 91 canonical subskills nested under their parent domains.

For each paper, the agent assigns one or more labels from the 14 capability domains and produces an evidence-strength label in \{strong, weak\} for each assigned domain, together with the set of endorsed subskills conditional on an assigned parent domain. A strong-evidence label is emitted when the paper explicitly names or operationalizes the capability as a central object of study and supports the claim through a dedicated task, benchmark, diagnostic experiment, mechanistic analysis, or primary conclusion. A weak-evidence label is emitted when the capability appears indirectly, e.g., as a supporting construct, task context, downstream behavior, or enabling mechanism. Domains are not treated as mutually exclusive: a paper may receive multiple domain and subskill labels when it substantively operationalizes overlapping constructs. Conceptual overlap among cognitive labels is permitted when supported by the text, but lexical overlap with engineering terminology alone is insufficient to license a cognitive capability label.

To keep the label space consistent across the corpus, subskill outputs are restricted to the canonical inventory of 91 subskills. During aggregation, capitalization and punctuation variants are normalized to canonical labels, unmatched labels are excluded from subskill-level analyses, and every subskill must attach to its designated parent capability. Language-Semantic Competence and Language-Pragmatic Competence are maintained as distinct domains, so that fine-grained separations in the label space are preserved rather than collapsed at annotation time.

\subsubsection{Annotation Pipeline}

The annotation agent is implemented as a multi-stage pipeline that takes a paper as input and produces the two label categories defined in Section 4.1.2. The pipeline has three components --- paper processing, structured output, and multi-model integration --- and is evaluated against a domain-expert reference set.

\textbf{Paper processing.} Each paper is first parsed into a structured text representation containing the sections most informative for capability annotation: abstract, introduction, methods, dataset or benchmark description, experimental setup, results, discussion, conclusion, and limitations. The agent then runs a relevance-screening stage; only papers satisfying the LLM-focused inclusion criterion proceed to capability annotation, while irrelevant papers receive annotation JSON objects with the complete schema populated with ``absence'' labels, preserving an auditable record over the full screening pool. Of the 31,505 screened papers, 15,934 were judged relevant and entered the capability analyses.

\textbf{Structured output.} For each retained paper, the agent produces a fixed-schema output consisting of (i) the final relevance judgment, (ii) assigned labels from the 14 domains and their corresponding domain-level evidence-strength labels, and (iii) the subskill sets conditional on assigned domains. Every output is subject to structural validation: capability labels must appear in the correct taxonomy layer; an "absence" label cannot co-occur with a substantive label within the same layer; and each subskill must correspond to its designated parent capability. Outputs that fail validation are re-emitted before entering the aggregation stage.

\textbf{Multi-model integration.} To reduce dependence on any single model's interpretation of ambiguous capability claims, three annotator models---DeepSeek V3.2, Kimi K2.6, and Claude Sonnet 4.6---independently run the pipeline over every paper. All three receive an identical prompt specifying the relevance criterion, the 14-domain taxonomy, the subskill inventory, the evidence-strength definitions, and the negative-mapping rules. Their structurally valid outputs then enter the consensus and arbitration procedure described in Section 4.1.4.

\textbf{Reliability.} To evaluate the pipeline against domain expertise, two cognitive-science experts independently reviewed a stratified random sample of 50 papers drawn from the full screening pool, covering both papers retained in and excluded from the analytic corpus. Using the same taxonomy and annotation guidelines, they independently assessed LLM-focused relevance and assigned capability-domain labels. Domain-level inter-rater agreement was Cohen's \(\kappa\) = 0.70.

% Alt text: Taxonomy-based literature annotation workflow from conference paper input through relevance screening, three-model extraction, consensus or arbitration, and final JSON output.
\begin{figure}[p]
\centering
\includegraphics[width=0.94\linewidth,height=0.82\textheight,keepaspectratio]{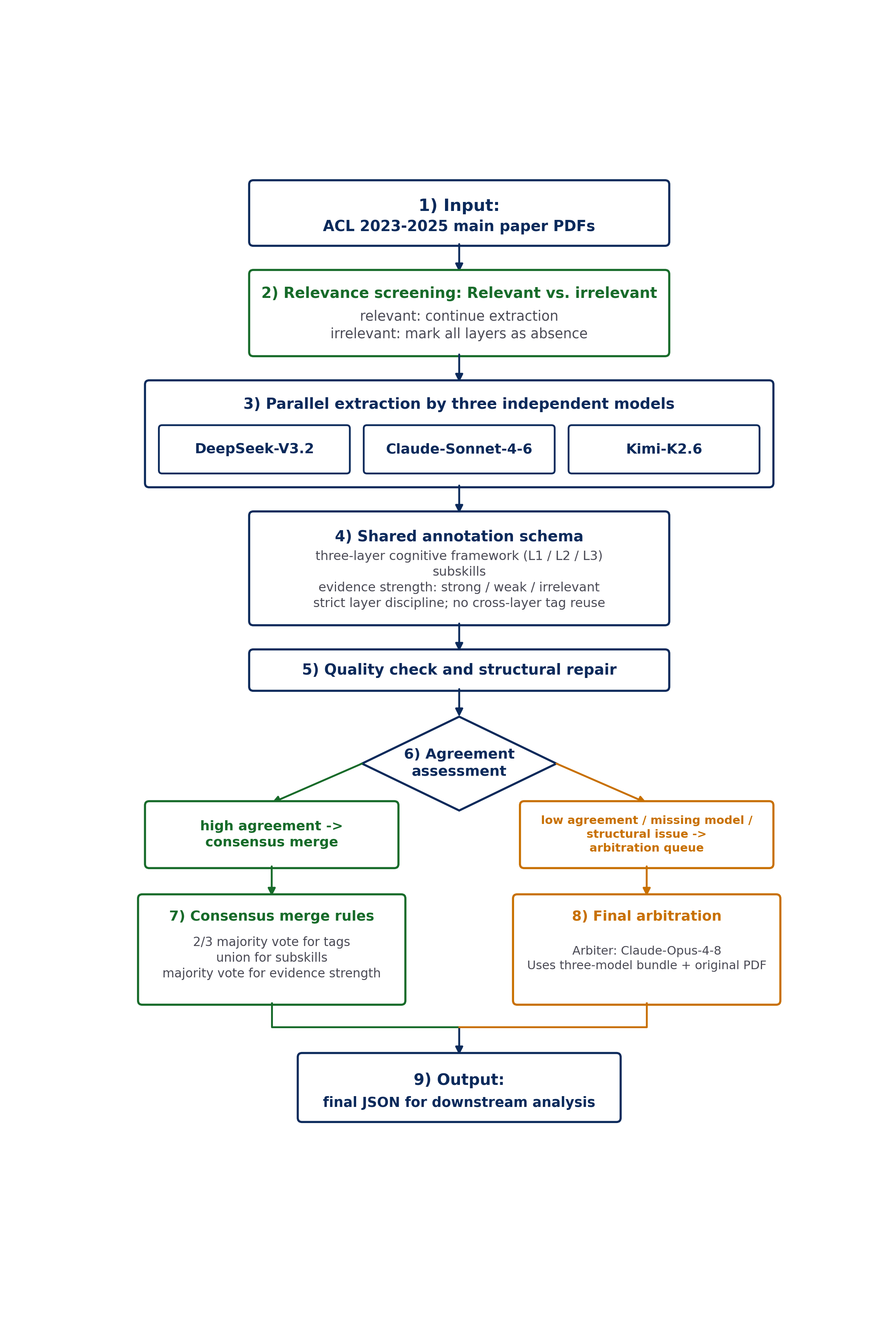}
\caption{Taxonomy-based literature annotation pipeline, with ACL shown as the input example; the same workflow ran on the AAAI, ACL, ICML, and NeurIPS screening pools.}
\label{fig:2}
\end{figure}

\subsubsection{Consensus and Arbitration}

The three annotations for each paper were first subjected to structural validation. Capability labels were required to appear in the correct taxonomy layer, an absent label could not co-occur with a substantive label within the same layer, and each subskill had to correspond to its designated parent capability. The final relevance judgment determined whether the paper entered the analytic corpus.

We then computed layer-level agreement across the three models' sets of non-absent capability labels, treating unanimous absence of capabilities as agreement. A paper was resolved by two-of-three majority voting when its mean layer-level agreement was at least 0.60 and no layer contained an unresolved tie. A capability label required support from at least two models. Subskills were pooled from models that endorsed the corresponding parent capability, and ties in evidence-strength ratings were resolved conservatively in favor of weak evidence.

Of the 31,505 screened papers, 22,922 were resolved directly through consensus. The remaining 8,583 papers were sent to Claude Opus 4.8 for arbitration. For each paper requiring arbitration, Claude Opus 4.8 received the original PDF, the three model annotations, and the complete taxonomy and annotation rules.

\subsubsection{Analysis Plan}

For \textbf{RQ1}, we estimated paper-level domain prevalence within the 15,934-paper analytic corpus. Strong-evidence labels served as the primary indicator of direct research attention, while weak-only labels were reported separately to capture supporting or contextual references.

For \textbf{RQ2}, we counted canonical subskills among papers assigned strong evidence for the corresponding conceptual domain. Percentages were calculated using the number of strong-evidence papers for that domain as the denominator. Because a paper could be assigned multiple subskills, these percentages are non-exclusive and may sum to more than 100\%. For consistency across domains, the main results table reports the three most frequently assigned subskills for each domain.

For \textbf{RQ3}, we analyzed paper-level co-occurrence between pairs of domains for which both labels received strong evidence. For each pair of domains (A) and (B), we computed the raw co-occurrence count (\emph{n\textsubscript{AB}}),

Jaccard similarity,

\[J(A,B)\  = \ \frac{n_{\text{AB}}}{n_{A}\  + \ n_{B}\  - \ n_{\text{AB}}}\]

Lift:

\[Lift(A,B)\  = \ \frac{\frac{n_{\text{AB}}}{N}}{(\frac{n_{A}}{N})(\frac{n_{B}}{N})}\]

where (\emph{n\textsubscript{A}}) and (\emph{n\textsubscript{B}}) denote the numbers of papers assigned strong evidence for domains (A) and (B), respectively, and (\emph{N} = 15,934). Jaccard similarity measures the degree of overlap between two domains, whereas lift indicates whether they co-occur more often than expected from their marginal prevalence. A lift greater than 1 indicates positive association, not functional dependency or causality. To reduce instability from rare combinations, normalized pairwise rankings were restricted to domain pairs that co-occurred in at least 20 papers.

Before aggregation, we deterministically re-applied the final relevance criterion so that papers classified as irrelevant contributed no capability labels. This post-merge validation identified and removed one contradictory record in which a nonempty capability annotation had been retained for an irrelevant paper. For the \textbf{RQ2} analysis, we additionally excluded 196 strong-evidence subskill assignments that did not match the finalized construct-label scheme.

\subsection{Results}

We organize the results around \textbf{RQ1-RQ3}, examining three complementary aspects of the literature: domain-level prevalence, within-domain subskill representation, and cross-domain co-occurrence.

\subsubsection{RQ1: Distribution of Cognitive Capability Domains}

Strong-evidence research attention was concentrated in a small number of domains (Table 3). \emph{Language-Semantic Competence} was directly studied in 3,551 papers (22.3\% of the analytic corpus), followed by \emph{Reasoning} (3,388; 21.3\%), \emph{Planning and Decision-Making} (2,149; 13.5\%), and \emph{Perception} (1,954; 12.3\%). \emph{Language-Pragmatic Competence} appeared as a direct research target in 560 papers (3.5\%). Each of the remaining domains appeared as a strong-evidence target in fewer than 6\% of papers.

At the layer level, at least one strong-evidence label was assigned to 3,239 papers for the Primitive layer (20.3\%), 5,294 for the Constructed layer (33.2\%), and 2,937 for the Integrative layer (18.4\%). Figure 3 visualizes paper-level strong-evidence prevalence for each domain, grouped by taxonomy layer. Because papers may receive multiple domain labels, the percentages are non-exclusive and do not sum to 100\%.

\begingroup
\scriptsize
\begin{longtable}[]{@{}
>{\raggedright\arraybackslash}p{(\linewidth-12\tabcolsep)*\real{0.070}}
>{\raggedright\arraybackslash}p{(\linewidth-12\tabcolsep)*\real{0.230}}
>{\raggedright\arraybackslash}p{(\linewidth-12\tabcolsep)*\real{0.100}}
>{\raggedright\arraybackslash}p{(\linewidth-12\tabcolsep)*\real{0.100}}
>{\raggedright\arraybackslash}p{(\linewidth-12\tabcolsep)*\real{0.140}}
>{\raggedright\arraybackslash}p{(\linewidth-12\tabcolsep)*\real{0.140}}
>{\raggedright\arraybackslash}p{(\linewidth-12\tabcolsep)*\real{0.120}}@{}}
\caption{Distribution of Strong and Weak-Only Evidence Across the 14 Domains}\tabularnewline
\toprule
\textbf{Layer} & \textbf{Domain} & \textbf{Strong n} & \textbf{Strong \%} & \textbf{Weak-only n} & \textbf{Weak-only \%} & \textbf{Mapped n} \\
\midrule
\endfirsthead
\toprule
\textbf{Layer} & \textbf{Domain} & \textbf{Strong n} & \textbf{Strong \%} & \textbf{Weak-only n} & \textbf{Weak-only \%} & \textbf{Mapped n} \\
\midrule
\endhead
L1 & Perception & 1,954 & 12.3\% & 4,546 & 28.5\% & 6,500 \\
& Attention & 722 & 4.5\% & 344 & 2.2\% & 1,066 \\
& Memory & 916 & 5.7\% & 475 & 3.0\% & 1,391 \\
L2 & Language-Semantic Competence & 3,551 & 22.3\% & 2,764 & 17.3\% & 6,315 \\
& Language-Pragmatic Competence & 560 & 3.5\% & 2,192 & 13.8\% & 2,752 \\
& Reasoning & 3,388 & 21.3\% & 1,126 & 7.1\% & 4,514 \\
& Emotion & 162 & 1.0\% & 35 & 0.2\% & 197 \\
& Theory of Mind & 114 & 0.7\% & 25 & 0.2\% & 139 \\
L3 & Metacognition & 482 & 3.0\% & 285 & 1.8\% & 767 \\
& Social Reasoning \& Interaction & 281 & 1.8\% & 80 & 0.5\% & 361 \\
& Planning \& Decision-Making & 2,149 & 13.5\% & 1,302 & 8.2\% & 3,451 \\
& Creativity \& Innovation & 75 & 0.5\% & 14 & 0.1\% & 89 \\
& Cultural Competence & 95 & 0.6\% & 34 & 0.2\% & 129 \\
& Moral Reasoning & 100 & 0.6\% & 34 & 0.2\% & 134 \\
\bottomrule
\end{longtable}
\endgroup

\emph{Note. Percentages use the 15,934 LLM-focused papers as the denominator. Strong and weak-only evidence are mutually exclusive within each domain at the paper level, and mapped (n) is their sum. Papers may be assigned to multiple domains; domain percentages are therefore non-exclusive and do not sum to 100\%.}

Weak-only labels reveal a complementary pattern. \emph{Perception} appeared as a supporting or contextual capability in 4,546 papers (28.5\%), compared with 1,954 papers in which it was a direct research target. \emph{Language-Semantic Competence} appeared as weak-only evidence in 2,764 papers (17.3\%), and \emph{Language-Pragmatic Competence} in 2,192 (13.8\%). The contrast was especially pronounced for \emph{Language-Pragmatic Competence}, which appeared far more often as contextual or supporting evidence than as a direct research target (560 papers; 3.5\%). \emph{Planning and Decision-Making} (1,302; 8.2\%) and \emph{Reasoning} (1,126; 7.1\%) were the next most frequent weak-only domains.

Six domains, \emph{Social Reasoning and Interaction}, \emph{Emotion}, \emph{Theory of Mind}, \emph{Creativity and Innovation}, \emph{Cultural Competence}, and \emph{Moral Reasoning}, each received strong-evidence coverage in fewer than 2\% of papers. RQ1 therefore reveals not only substantial inequality in domain-level coverage, but also differences in how capabilities enter the literature: some are studied primarily as direct targets, whereas others appear more often as indirect or contextual components.

% Alt text: Bar chart of paper-level strong-evidence prevalence for the 14 cognitive capability domains, grouped by taxonomy layer.
\begin{figure}[p]
\centering
\includegraphics[width=0.94\linewidth,height=0.82\textheight,keepaspectratio]{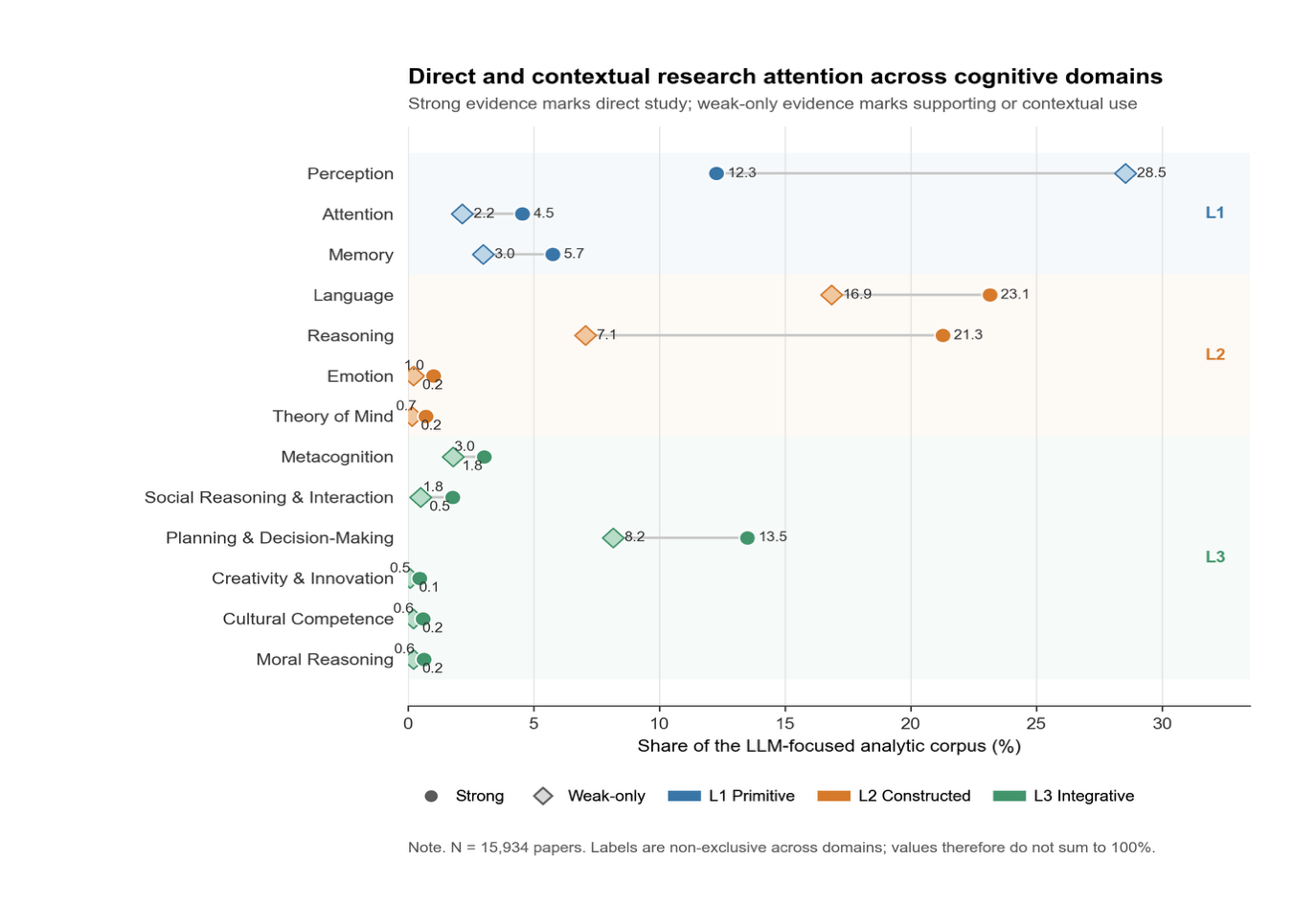}
\caption{Paper-level prevalence of strong-evidence cognitive capability domains across the three taxonomy layers. Bars show the percentage of the 15,934-paper analytic corpus assigned strong evidence for each domain. Domains are grouped by taxonomy layer. Because papers may receive multiple domain labels, percentages are non-exclusive and do not sum to 100\%.}
\label{fig:3}
\end{figure}

\subsubsection{RQ2: Fine-Grained Representation of Capability Subskills}

Subskill representation varied substantially within the 14 domains (Table 4). The most frequent subskill appeared in a median of 97.9\% of strong-evidence papers within its domain, with domain-level values ranging from 67.7\% to 100.0\%. In 10 of the 14 domains, the most frequent subskill appeared in at least 90\% of strong-evidence papers, and in 12 domains it appeared in at least 80\%. Table 4 reports the three most frequently represented subskills in each domain. Because papers could receive multiple subskill labels, the percentages indicate prevalence within a domain rather than mutually exclusive shares.

\begingroup
\scriptsize
\begin{longtable}[]{@{}
>{\raggedright\arraybackslash}p{(\linewidth-10\tabcolsep)*\real{0.105}}
>{\raggedright\arraybackslash}p{(\linewidth-10\tabcolsep)*\real{0.235}}
>{\raggedright\arraybackslash}p{(\linewidth-10\tabcolsep)*\real{0.120}}
>{\raggedright\arraybackslash}p{(\linewidth-10\tabcolsep)*\real{0.180}}
>{\raggedright\arraybackslash}p{(\linewidth-10\tabcolsep)*\real{0.180}}
>{\raggedright\arraybackslash}p{(\linewidth-10\tabcolsep)*\real{0.180}}@{}}
\caption{Three Most Frequent Subskills Within Each Strong-Evidence Domain}\tabularnewline
\toprule
\textbf{Layer} & \textbf{Domain} & \textbf{Strong n} & \textbf{Subskill 1} & \textbf{Subskill 2} & \textbf{Subskill 3} \\
\midrule
\endfirsthead
\toprule
\textbf{Layer} & \textbf{Domain} & \textbf{Strong n} & \textbf{Subskill 1} & \textbf{Subskill 2} & \textbf{Subskill 3} \\
\midrule
\endhead
L1 & Perception & 1,954 & Feature extraction \& pattern recognition: 1,938 (99.2\%) & Multisensory integration \& cue combination: 1,387 (71.0\%) & Perceptual organization \& segmentation: 881 (45.1\%) \\
L1 & Attention & 722 & Selective attention: 717 (99.3\%) & Sustained attention: 158 (21.9\%) & Orienting attention: 65 (9.0\%) \\
L1 & Memory & 916 & Working memory: 633 (69.1\%) & Episodic memory: 377 (41.2\%) & Procedural memory: 245 (26.7\%) \\
L2 & Language-Semantic Competence & 3,551 & Compositional semantics: 3,245 (91.4\%) & Lexical-semantic knowledge: 3,081 (86.8\%) & Discourse-level meaning: 3,003 (84.6\%) \\
L2 & Language-Pragmatic Competence & 560 & Contextual appropriateness: 546 (97.5\%) & Implicature inference: 377 (67.3\%) & Sociocultural meaning: 193 (34.5\%) \\
L2 & Reasoning & 3,388 & Deductive reasoning: 2,292 (67.7\%) & Inductive reasoning: 2,136 (63.0\%) & Analogical and relational reasoning: 1,842 (54.4\%) \\
L2 & Emotion & 162 & Emotion perception and recognition: 161 (99.4\%) & Emotion understanding and clarity: 105 (64.8\%) & Emotion utilization: 27 (16.7\%) \\
L2 & Theory of Mind & 114 & Perspective taking and percepts: 94 (82.5\%) & ToM of intention: 92 (80.7\%) & ToM of beliefs: 75 (65.8\%) \\
L3 & Metacognition & 482 & Metacognitive monitoring: 474 (98.3\%) & Metacognitive regulation: 275 (57.1\%) & Metacognitive knowledge: 82 (17.0\%) \\
L3 & Social Reasoning \& Interaction & 281 & Contextual integration of social information: 230 (81.9\%) & Social appropriateness judgment: 217 (77.2\%) & Other-perspective inference: 141 (50.2\%) \\
L3 & Planning \& Decision-Making & 2,149 & Evaluation, comparison, and selection (decision rule use): 1,999 (93.0\%) & Option generation and alternative construction: 1,672 (77.8\%) & Problem framing and goal specification: 1,635 (76.1\%) \\
L3 & Creativity \& Innovation & 75 & Creative ideation capacity: 75 (100.0\%) & Cognitive flexibility (creative thinking style): 59 (78.7\%) & Boundary-pushing problem solving: 35 (46.7\%) \\
L3 & Cultural Competence & 95 & Contextual cultural knowledge integration: 95 (100.0\%) & Culturally responsive practice adaptation: 60 (63.2\%) & Adaptive communication: 59 (62.1\%) \\
L3 & Moral Reasoning & 100 & Moral judgment (cognitive evaluation): 100 (100.0\%) & Moral motivation (integrity/\allowbreak{}commitment): 20 (20.0\%) & Moral implementation (courage and action): 11 (11.0\%) \\
\bottomrule
\end{longtable}
\endgroup

\emph{Note. Percentages use each domain's strong-evidence count as the denominator. Subskill labels follow the final construct-label scheme, with capitalization and punctuation variants normalized. Because papers could receive multiple subskill labels, percentages are non-exclusive. The same top-three reporting rule was applied to all domains.}

Two broad patterns emerged. Several domains were concentrated around one highly prevalent subskill, followed by substantially lower coverage of other subskills. In \emph{Attention}, selective attention appeared in 99.3\% of strong-evidence papers, compared with 21.9\% for sustained attention and 9.0\% for orienting attention. Similar declines were observed in \emph{Metacognition}, from 98.3\% for metacognitive monitoring to 57.1\% for metacognitive regulation and 17.0\% for metacognitive knowledge, and in \emph{Moral Reasoning}, from 100.0\% for moral judgment to 20.0\% for moral motivation and 11.0\% for moral implementation.

Other domains showed broader co-coverage across several leading subskills. In \emph{Language-Semantic Competence}, compositional semantics, lexical-semantic knowledge, and discourse-level meaning each appeared in more than 84\% of strong-evidence papers. \emph{Reasoning} was comparatively distributed across deductive, inductive, and analogical or relational reasoning, with prevalence ranging from 54.4\% to 67.7\%. \emph{Planning and Decision-Making} also showed substantial coverage of evaluation and selection, option generation, and problem framing, each appearing in more than 76\% of papers.

The two language domains differed not only in overall representation but also in their internal subskill profiles. Semantic research commonly addressed several major subskills concurrently, whereas pragmatic research was more concentrated around contextual appropriateness (97.5\%), followed by implicature inference (67.3\%) and sociocultural meaning (34.5\%).

Overall, RQ2 shows that domain-level coverage concealed distinct internal profiles: some domains were represented primarily through a narrow subset of subskills, whereas others showed broader co-coverage among several leading subskills.

\subsubsection{RQ3: Strong-Evidence Capability Co-occurrence}

Capability co-occurrence revealed two complementary patterns: a high-volume structure centered on \emph{Language-Semantic Competence} and \emph{Reasoning}, and a lower-volume but strongly associated set of pragmatics, social, cultural, and moral capabilities. We report both raw co-occurrence frequency and normalized association because common domain pairs may occur frequently simply because each domain is individually prevalent, whereas lift identifies pairs that appear together more often than expected under statistical independence. Figures 4 and 5 provide an overview of both measures across domain pairs, and Table 5 reports the most frequent pairs and the strongest normalized associations.

The highest-volume pair was \emph{Language-Semantic Competence} and \emph{Reasoning}, which co-occurred in 1,864 papers, or 11.7\% of the analytic corpus. Other frequent combinations linked \emph{Perception} with \emph{Language-Semantic Competence} (n = 720, 4.5\%), \emph{Reasoning} with \emph{Planning and Decision-Making} (n = 698, 4.4\%), \emph{Perception} with \emph{Reasoning} (n = 557, 3.5\%), and \emph{Language-Semantic Competence} with \emph{Planning and Decision-Making} (n = 446, 2.8\%). These results indicate that the most visible joint research targets were concentrated around semantic language, reasoning, perception, and decision-related capabilities.

Raw frequency, however, did not always indicate a distinctive association. \emph{Language-Semantic Competence} and \emph{Reasoning} were both frequently co-studied and positively associated relative to their marginal prevalence (n = 1,864; lift = 2.47). By contrast, \emph{Language-Semantic Competence} and \emph{Planning and Decision-Making} had a relatively high co-occurrence count but a lift slightly below 1 (n = 446; lift = 0.93), suggesting that their joint frequency was largely attributable to the prevalence of the two individual domains. The two language domains showed the reverse pattern: \emph{Language-Semantic Competence} and \emph{Language-Pragmatic Competence} co-occurred in fewer papers (n = 424, 2.7\%) but were studied together 3.40 times as often as expected under independence.

The strongest normalized associations occurred among less prevalent social, pragmatic, cultural, and moral domains. \emph{Theory of Mind} and \emph{Social Reasoning and Interaction} showed the highest lift among pairs meeting the minimum frequency threshold (n = 62; lift = 30.84). Other strong associations linked \emph{Social Reasoning and Interaction} with \emph{Cultural Competence} (n = 22; lift = 13.13), \emph{Language-Pragmatic Competence} with \emph{Social Reasoning and Interaction} (n = 124; lift = 12.56), \emph{Social Reasoning and Interaction} with \emph{Moral Reasoning} (n = 22; lift = 12.48), and \emph{Language-Pragmatic Competence} with \emph{Cultural Competence} (n = 40; lift = 11.98). Although uncommon in absolute terms, these domains were repeatedly studied together when they appeared, indicating a comparatively cohesive but low-volume strand of research attention. Associations close to the minimum frequency threshold should nevertheless be interpreted as exploratory.

Overall, RQ3 identifies a contrast between research volume and relational specificity. High-volume co-occurrence was dominated by \emph{Language-Semantic Competence}, \emph{Reasoning}, \emph{Perception}, and \emph{Planning and Decision-Making}, whereas the strongest relative associations centered on \emph{Social Reasoning and Interaction}, linking it to \emph{Theory of Mind}, \emph{Cultural Competence}, \emph{Language-Pragmatic Competence}, and \emph{Moral Reasoning}. A further strong association connected \emph{Language-Pragmatic Competence} with \emph{Cultural Competence}. These patterns characterize how capability domains are jointly studied; they do not establish causal direction or functional dependency.

% Alt text: Bubble matrix of pairwise domain co-occurrence counts and lift among papers with strong evidence for both domains.
\begin{figure}[p]
\centering
\includegraphics[width=0.94\linewidth,height=0.82\textheight,keepaspectratio]{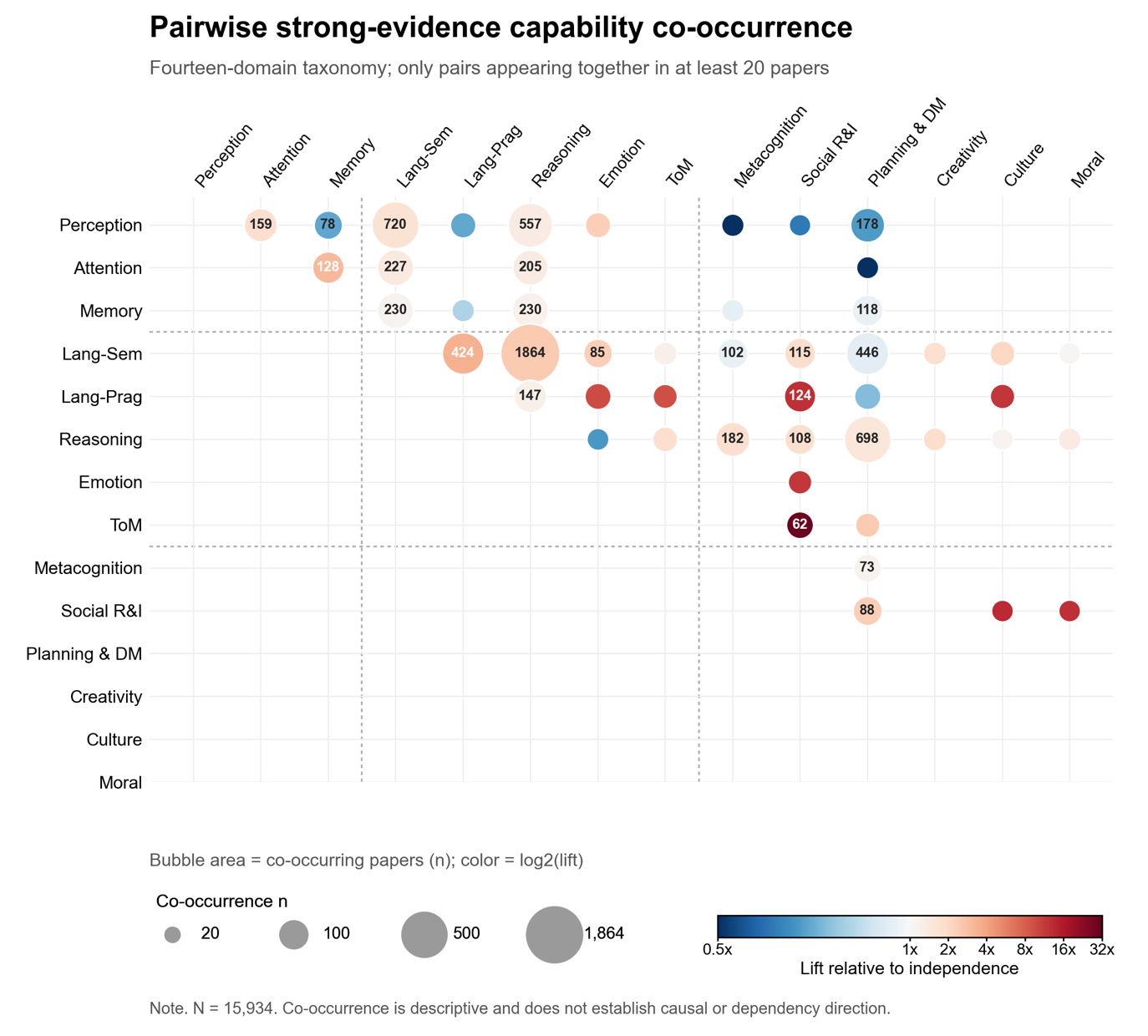}
\caption{Pairwise co-occurrence frequency and normalized association across cognitive capability domains. Bubble area represents the number of papers with strong evidence for both domains, and color represents lift relative to statistical independence. Only pairs with at least 20 co-occurrences are shown.}
\label{fig:4}
\end{figure}

% Alt text: Positive-association network in which node size represents domain prevalence and edges represent co-occurrence pairs with lift greater than one.
\begin{figure}[p]
\centering
\includegraphics[width=0.94\linewidth,height=0.82\textheight,keepaspectratio]{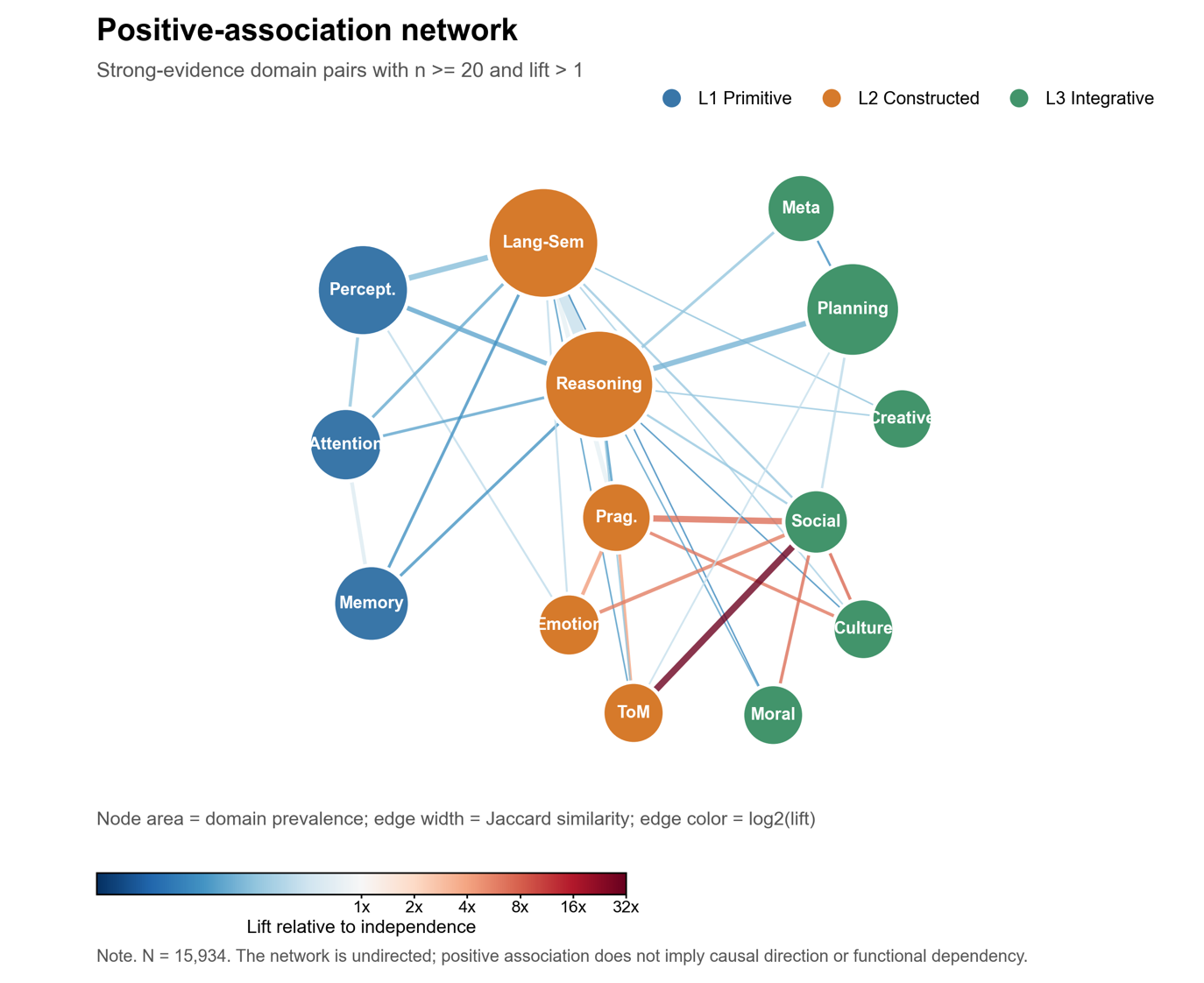}
\caption{Positive-association network of cognitive capability domains. Nodes represent domains and are scaled by strong-evidence prevalence. Edges are restricted to pairs with at least 20 co-occurrences and lift greater than 1; edge width represents Jaccard similarity and edge color represents lift. The network characterizes the organization of research attention rather than causal or functional dependency.}
\label{fig:5}
\end{figure}

\begingroup
\scriptsize
\begin{longtable}[]{@{}
>{\raggedright\arraybackslash}p{(\linewidth-10\tabcolsep)*\real{0.105}}
>{\raggedright\arraybackslash}p{(\linewidth-10\tabcolsep)*\real{0.235}}
>{\raggedright\arraybackslash}p{(\linewidth-10\tabcolsep)*\real{0.120}}
>{\raggedright\arraybackslash}p{(\linewidth-10\tabcolsep)*\real{0.180}}
>{\raggedright\arraybackslash}p{(\linewidth-10\tabcolsep)*\real{0.180}}
>{\raggedright\arraybackslash}p{(\linewidth-10\tabcolsep)*\real{0.180}}@{}}
\caption{Most Frequent and Strongest Normalized Capability Co-occurrences}\tabularnewline
\toprule
\textbf{Selection criterion} & \textbf{Capability pair} & \textbf{Co-occurrence n} & \textbf{\% of corpus} & \textbf{Jaccard} & \textbf{Lift} \\
\midrule
\endfirsthead
\toprule
\textbf{Selection criterion} & \textbf{Capability pair} & \textbf{Co-occurrence n} & \textbf{\% of corpus} & \textbf{Jaccard} & \textbf{Lift} \\
\midrule
\endhead
Highest frequency & Language-Semantic Competence + Reasoning & 1,864 & 11.7\% & 0.367 & 2.47 \\
& Perception + Language-Semantic Competence & 720 & 4.5\% & 0.150 & 1.65 \\
& Reasoning + Planning \& Decision-Making & 698 & 4.4\% & 0.144 & 1.53 \\
& Perception + Reasoning & 557 & 3.5\% & 0.116 & 1.34 \\
& Language-Semantic Competence + Planning \& Decision-Making & 446 & 2.8\% & 0.085 & 0.93 \\
& Language-Semantic Competence + Language-Pragmatic Competence & 424 & 2.7\% & 0.115 & 3.40 \\
& Memory + Language-Semantic Competence & 230 & 1.4\% & 0.054 & 1.13 \\
& Memory + Reasoning & 230 & 1.4\% & 0.056 & 1.18 \\
Highest lift (n \textgreater= 20) & Theory of Mind + Social Reasoning \& Interaction & 62 & 0.4\% & 0.186 & 30.84 \\
& Social Reasoning \& Interaction + Cultural Competence & 22 & 0.1\% & 0.062 & 13.13 \\
& Language-Pragmatic Competence + Social Reasoning \& Interaction & 124 & 0.8\% & 0.173 & 12.56 \\
& Social Reasoning \& Interaction + Moral Reasoning & 22 & 0.1\% & 0.061 & 12.48 \\
& Language-Pragmatic Competence + Cultural Competence & 40 & 0.3\% & 0.065 & 11.98 \\
\bottomrule
\end{longtable}
\endgroup

\emph{Note. Co-occurrence required strong evidence for both domains in the same paper. The frequency panel reports the eight largest raw counts. The normalized panel reports the five highest lift values among pairs not already included in the frequency panel and with at least 20 co-occurring papers. Jaccard measures proportional overlap, whereas lift compares observed co-occurrence with that expected under independence. Estimates near the minimum count threshold should be interpreted cautiously.}

\section{Discussion and Conclusion}

This study introduces a multi-layer taxonomy that organizes LLM cognitive capabilities into 14 domains, 91 subskills, and three layers. The literature mapping demonstrates what this capability-level representation reveals beyond task- and benchmark-based classifications. Three findings stand out. First, research attention is concentrated on \emph{Language-Semantic Competence}, \emph{Reasoning}, \emph{Planning and Decision-Making}, and \emph{Perception}, whereas six domains appear in fewer than 2\% of the analyzed papers. Second, nominal domain coverage often masks substantial subskill concentration: the most frequently studied subskill accounts for a median of 97.9\% of papers within its domain and exceeds 90\% in 10 of the 14 domains. Third, cross-domain research exhibits both a high-volume core centered on \emph{Language-Semantic Competence} and \emph{Reasoning} and a lower-volume cluster of strong associations among pragmatic, social, cultural, and moral capabilities.

Together, these findings show that the research landscape is uneven at three levels: which domains receive attention, which subskills represent those domains, and which capabilities are studied together. The analysis provides an initial demonstration of the taxonomy as a framework for research mapping and motivates its broader use in benchmark design, capability diagnosis, and model-development research.

\subsection{From Task Categories to a Structured Capability Space}

The taxonomy shifts the unit of analysis from evaluation artifacts to the capabilities they are intended to probe. Tasks, benchmarks, datasets, and deployment scenarios provide important evidence, but they do not by themselves offer a common representation of the constructs being evaluated. A single task may recruit multiple capabilities, while the same capability may be expressed across different tasks and modalities. Capability-level annotation therefore enables comparison across otherwise disconnected benchmark families and application domains.

The taxonomy extends this approach in two ways. First, its 91-subskill codebook distinguishes broad domain coverage from the specific forms of capability examined within each domain. As the mapping results show, a domain can appear well represented even when most studies concentrate on a single subskill. Second, the three-layer organization encodes hypothesized support relations rather than treating all capabilities as independent entries in a flat list. This structure allows researchers to distinguish the capability targeted by an evaluation from candidate supporting capabilities that may also affect performance.

The taxonomy thus provides a common analytical layer for integrating heterogeneous evaluation evidence, auditing coverage at both domain and subskill levels, and formulating testable hypotheses about relations among capabilities.

\subsection{Implications for Evaluation Design and Capability Diagnosis}

The taxonomy supports evaluation at three stages: construct specification, benchmark composition, and performance interpretation.

At the specification stage, evaluations should identify the capability domains and subskills they target, in addition to their task formats and application settings. Similar tasks may probe different capabilities, while the same capability may be expressed across different tasks and modalities. Explicit capability labels therefore make evaluation objectives easier to compare.

At the composition stage, benchmark suites can be audited as capability portfolios. Developers can assess which domains and subskills are covered, where coverage is missing, and whether multiple benchmarks repeatedly measure similar capability profiles. The goal is not uniform coverage across all domains, but coverage that is explicit and appropriate for the evaluation's intended use.

At the interpretation stage, the multi-layer structure supports more targeted diagnosis. Poor performance on an Integrative capability may reflect limitations in the target itself or in candidate supporting capabilities. Follow-up probes can distinguish between these explanations and identify potential lower-layer bottlenecks. These analyses generate structured diagnostic hypotheses; they do not by themselves establish causal dependencies.

The taxonomy thus complements existing benchmarks by placing heterogeneous evaluation tasks within a common capability representation. It clarifies what each benchmark targets, which supporting capabilities may affect performance, and which regions of the capability space remain unmeasured. This representation can improve the interpretation of both individual benchmark results and aggregate model profiles.

\subsection{Implications for Training and Transfer Research}

The taxonomy defines a hypothesis space for capability-targeted training. Researchers can compare direct training on an Integrative capability with staged training that first targets candidate supporting capabilities. These experiments can test whether layer-guided sequencing improves sample efficiency, robustness, or generalization.

The 91-subskill codebook also provides a structured basis for data selection and transfer analysis. Training examples can be annotated by the capabilities they engage, enabling comparisons across tasks that share a target capability, overlap in candidate supports, or have distinct capability profiles. This design helps separate task-specific adaptation, within-capability generalization, and cross-capability transfer.

Candidate support relations require causal testing. Controlled interventions, ablations, and matched-task comparisons can test whether improving a lower-layer capability reliably benefits a higher-layer target and whether gains arise from capability transfer rather than shared surface features, benchmark formats, or training data. The literature mapping can prioritize understudied domains and capability combinations, but it cannot determine training order or transfer direction. Future work should test whether layer-guided interventions reduce ability inversion, improve transfer, and distinguish target-specific failures from lower-layer bottlenecks.

\subsection{Limitations and Boundary Conditions}

This study has several limitations. First, the taxonomy is not a blueprint for LLM architecture. The proposed domains, layers, and capability relations should therefore be viewed as conceptual and diagnostic hypotheses, not as established properties of LLMs.

Second, the literature mapping operates at the paper level and relies on construct annotation rather than direct model evaluation. Although the annotation pipeline included multi-model annotation, arbitration, and expert review, some ambiguity remains for conceptually adjacent domains. Moreover, because the taxonomy permits non-exclusive labels, co-occurrence patterns should be interpreted as descriptions of research organization rather than evidence of causal or functional dependency.

Third, the corpus includes LLM-focused papers from ACL, AAAI, ICML, and NeurIPS between 2023 and 2025. The reported patterns therefore characterize research attention within the selected venues and period rather than the entire LLM capability literature.

\subsection{Conclusion}

This paper introduced a multi-layer taxonomy that organizes LLM cognitive capabilities into 14 domains, 91 subskills, and three hierarchical layers. Applying the taxonomy to recent LLM research demonstrated how a capability-centered representation can reveal systematic patterns in coverage and cross-capability organization that remain obscured by task- and benchmark-based classifications. Rather than serving as a new benchmark or a mechanistic model of LLM cognition, the taxonomy provides a common representational framework for capability specification, coverage analysis, diagnostic hypothesis generation, and future empirical research on evaluation, training, and transfer.

\section*{References}
Albert, D., \& Steinberg, L. (2011). Age differences in strategic planning as indexed by the Tower of London. Child Development, 82(5), 1501-1517. https://doi.org/10.1111/j.1467-8624.2011.01613.x

Baddeley, A. (2003). Working memory and language: An overview. Journal of Communication Disorders, 36(3), 189-208. https://doi.org/10.1016/S0021-9924(03)00019-4

Badre, D., \& Nee, D. E. (2018). Frontal cortex and the hierarchical control of behavior. Trends in Cognitive Sciences, 22(2), 170-188. https://doi.org/10.1016/j.tics.2017.11.005

Banich, M. T., \& Compton, R. J. (2023). Cognitive neuroscience (5th ed.). Cambridge University Press.

Bean, A. M., Kearns, R. O., Romanou, A., Hafner, F. S., Mayne, H., Batzner, J., Foroutan Eghlidi, N., Schmitz, C., Korgul, K., Batra, H., Deb, O., Beharry, E., Emde, C., Foster, T., Gausen, A., Grandury, M., Han, S., Hofmann, V., Ibrahim, L., Kim, H., Kirk, H. R., Lin, F., Liu, G., Luettgau, L., Magomere, J., Rystrøm, J., Sotnikova, A., Yang, Y., Zhao, Y., Bibi, A., Bosselut, A., Clark, R., Cohan, A., Foerster, J., Gal, Y., Hale, S., Raji, D., Summerfield, C., Torr, P., Ududec, C., Rocher, L., \& Mahdi, A. (2025). Measuring what matters: Construct validity in large language model benchmarks. Advances in Neural Information Processing Systems, 38.

Bisk, Y., Holtzman, A., Thomason, J., Andreas, J., Bengio, Y., Chai, J., Lapata, M., Lazaridou, A., May, J., Nisnevich, A., Pinto, N., \& Turian, J. (2020). Experience grounds language. In Proceedings of the 2020 Conference on Empirical Methods in Natural Language Processing (pp.~8718-8735). Association for Computational Linguistics. https://doi.org/10.18653/v1/2020.emnlp-main.703

Burnell, R., Yamamori, Y., Firat, O., Olszewska, K., Hughes-Fitt, S., Kelly, O., Galatzer-Levy, I. R., Morris, M. R., Dafoe, A., Snyder, A. M., Goodman, N. D., Botvinick, M., \& Legg, S. (2026). Measuring progress toward AGI: A cognitive framework. arXiv preprint arXiv:2605.28405.

Chang, Y., Wang, X., Wang, J., Wu, Y., Yang, L., Zhu, K., Chen, H., Yi, X., Wang, C., Wang, Y., Ye, W., Zhang, Y., Chang, Y., Yu, P. S., Yang, Q., \& Xie, X. (2024). A survey on evaluation of large language models. ACM Transactions on Intelligent Systems and Technology, 15(3), Article 39, 1-45. https://doi.org/10.1145/3641289

Chen, M. F., Roberts, N., Bhatia, K., Wang, J., Zhang, C., Sala, F., \& Ré, C. (2023). Skill-it! A data-driven skills framework for understanding and training language models. Advances in Neural Information Processing Systems, 36.

Coda-Forno, J., Binz, M., Wang, J. X., \& Schulz, E. (2024). CogBench: A large language model walks into a psychology lab. arXiv preprint arXiv:2402.18225.

de Langis, K., Park, J. I., Hu, B., Le, K. C., Schramm, A., Mensink, M. C., Elfenbein, A., \& Kang, D. (2025). A framework for robust cognitive evaluation of LLMs. arXiv preprint arXiv:2504.02789.

Diamond, A. (2013). Executive functions. Annual Review of Psychology, 64, 135-168. https://doi.org/10.1146/annurev-psych-113011-143750

Eysenck, M. W., \& Keane, M. T. (2020). Cognitive psychology: A student's handbook (8th ed.). Psychology Press.

Fadda, R., Parisi, M., Ferretti, L., Saba, G., Foscoliano, M., Salvago, A., \& Doneddu, G. (2016). Exploring the role of theory of mind in moral judgment: The case of children with autism spectrum disorder. Frontiers in Psychology, 7, 523. https://doi.org/10.3389/fpsyg.2016.00523

Fischer, K. W. (1980). A theory of cognitive development: The control and construction of hierarchies of skills. Psychological Review, 87(6), 477-531. https://doi.org/10.1037/0033-295X.87.6.477

Gauvain, M. (2022). Cognitive development in infancy and childhood. Cambridge University Press. https://doi.org/10.1017/9781108955676

Gogtay, N., Giedd, J. N., Lusk, L., Hayashi, K. M., Greenstein, D., Vaituzis, A. C., Nugent, T. F., III, Herman, D. H., Clasen, L. S., Toga, A. W., Rapoport, J. L., \& Thompson, P. M. (2004). Dynamic mapping of human cortical development during childhood through early adulthood. Proceedings of the National Academy of Sciences, 101(21), 8174-8179.

Haznitrama, F. G., Ardi, F. R., \& Oh, A. (2026). A neuropsychologically grounded evaluation of LLM cognitive abilities. arXiv preprint arXiv:2603.02540.

Houdé, O., \& Borst, G. (2022). Cognitive development studies: From the history of psychology to the current trends in cognitive sciences. In O. Houdé \& G. Borst (Eds.), The Cambridge handbook of cognitive development (pp.~1-12). Cambridge University Press. https://doi.org/10.1017/9781108399838.001

Jain, N., et al.~(2025). LiveCodeBench: Holistic and contamination free evaluation of large language models for code. In Proceedings of the Thirteenth International Conference on Learning Representations.

Just, M. A., \& Carpenter, P. A. (1992). A capacity theory of comprehension: Individual differences in working memory. Psychological Review, 99(1), 122-149. https://doi.org/10.1037/0033-295X.99.1.122

Liang, P., Bommasani, R., Lee, T., Tsipras, D., Soylu, D., Yasunaga, M., Zhang, Y., Narayanan, D., Wu, Y., Kumar, A., et al.~(2023). Holistic evaluation of language models. Transactions on Machine Learning Research.

McCoy, R. T., Yao, S., Friedman, D., Hardy, M. D., \& Griffiths, T. L. (2024). Embers of autoregression show how large language models are shaped by the problem they are trained to solve. Proceedings of the National Academy of Sciences, 121(41), e2322420121. https://doi.org/10.1073/pnas.2322420121

Miller, E. K., \& Cohen, J. D. (2001). An integrative theory of prefrontal cortex function. Annual Review of Neuroscience, 24, 167-202. https://doi.org/10.1146/annurev.neuro.24.1.167

Mo, H., Ma, X., Liu, X., Wong, D. F., Li, Y., Liu, J., \& Zhang, M. (2025). CDT: A comprehensive capability framework for large language models across cognition, domain, and task. In Findings of the Association for Computational Linguistics: EMNLP 2025 (pp.~3715-3734).

Moran, J. M., Young, L. L., Saxe, R., Lee, S. M., O'Young, D., Mavros, P. L., \& Gabrieli, J. D. E. (2011). Impaired theory of mind for moral judgment in high-functioning autism. Proceedings of the National Academy of Sciences, 108(7), 2688-2692. https://doi.org/10.1073/pnas.1011734108

OpenAI. (2025). GPT-4.5 system card. https://openai.com/index/gpt-4-5-system-card/

Phan, L., et al.~(2025). Humanity's Last Exam. arXiv preprint arXiv:2501.14249.

Piaget, J. (1952). The origins of intelligence in children. International Universities Press.

Reisberg, D. (2013). Cognition: Exploring the science of the mind (5th ed.). W. W. Norton.

Reynolds, G. D., \& Richards, J. E. (2019). Infant visual attention and stimulus repetition effects on object recognition. Child Development, 90(4), 1023-1041. https://doi.org/10.1111/cdev.12982

Vygotsky, L. S. (1978). Mind in society: The development of higher psychological processes. Harvard University Press.

Wang, X., Yuan, P., Feng, S., Li, Y., Pan, B., Wang, H., Hu, Y., \& Li, K. (2024). CogLM: Tracking cognitive development of large language models. arXiv preprint arXiv:2408.09150.

Wei, J., Sun, Z., Papay, S., McKinney, S., Han, J., Fulford, I., Chung, H. W., Passos, A. T., Fedus, W., \& Glaese, A. (2025). BrowseComp: A simple yet challenging benchmark for browsing agents. arXiv preprint arXiv:2504.12516.

Wei, J., Wang, X., Schuurmans, D., Bosma, M., Ichter, B., Xia, F., Chi, E. H., Le, Q. V., \& Zhou, D. (2022). Chain-of-thought prompting elicits reasoning in large language models. Advances in Neural Information Processing Systems, 35, 24824-24837.

Wu, Z., Qiu, L., Ross, A., Akyürek, E., Chen, B., Wang, B., Kim, N., Andreas, J., \& Kim, Y. (2024). Reasoning or reciting? Exploring the capabilities and limitations of language models through counterfactual tasks. In Proceedings of the 2024 Conference of the North American Chapter of the Association for Computational Linguistics: Human Language Technologies (Volume 1: Long Papers) (pp.~1819-1862). Association for Computational Linguistics. https://doi.org/10.18653/v1/2024.naacl-long.102

Ye, H., Jin, J., Xie, Y., Zhang, X., \& Song, G. (2025). Large language model psychometrics: A systematic review of evaluation, validation, and enhancement. arXiv preprint arXiv:2505.08245.

Zhang, G., Ying, Y., Yue, G., Jiang, S., Liang, J., Fu, Y., Hu, H., \& Xiao, Y. (2025). From remembering to metacognition: Do existing benchmarks accurately evaluate LLMs? In Findings of the Association for Computational Linguistics: EMNLP 2025 (pp.~13440-13457).

\appendix
\setcounter{table}{0}
\renewcommand{\thetable}{A\arabic{table}}
\renewcommand{\theHtable}{A.\arabic{table}}
\section{Complete Construct-Label Scheme}

Appendix A presents the complete construct-label scheme used in the literature-mapping analysis. The scheme comprises 14 capability domains and 91 canonical subskills organized across the Primitive, Constructed, and Integrative layers. Domains are not mutually exclusive, and a paper may receive multiple labels when it substantively targets overlapping capability constructs. Each subskill retains a fixed parent-domain assignment for annotation and aggregation. Capitalization and punctuation have been standardized for presentation without changing construct content or parent-domain assignment. Table A1 summarizes the domain-level structure and subskill counts.

For labels that originate in human constructs involving subjective experience, motivation, or internal mechanisms, annotation is based only on observable model outputs, behavioral manifestations, or functional system-level analogues. Inclusion of such labels does not imply that an LLM possesses the corresponding human experience or mechanism.

\subsection{Taxonomy Overview}

\begingroup
\footnotesize
\begin{longtable}[]{@{}
>{\raggedright\arraybackslash}p{(\linewidth-6\tabcolsep)*\real{0.140}}
>{\raggedright\arraybackslash}p{(\linewidth-6\tabcolsep)*\real{0.220}}
>{\raggedright\arraybackslash}p{(\linewidth-6\tabcolsep)*\real{0.190}}
>{\raggedright\arraybackslash}p{(\linewidth-6\tabcolsep)*\real{0.450}}@{}}
\caption{Taxonomy Overview}\tabularnewline
\toprule
\textbf{Layer} & \textbf{Domain} & \textbf{Subskills} & \textbf{Domain-level definition} \\
\midrule
\endfirsthead
\toprule
\textbf{Layer} & \textbf{Domain} & \textbf{Subskills} & \textbf{Domain-level definition} \\
\midrule
\endhead
Primitive & \emph{Perception} & 5 & The functional encoding, extraction, integration, organization, recognition, and adaptive interpretation of structure in linguistic, symbolic, sensory, or multimodal input. \\
Primitive & \emph{Attention} & 3 & The orienting, selection, and sustained prioritization of task-relevant information while resisting irrelevant, distracting, or competing input. \\
Primitive & \emph{Memory} & 4 & The encoding, temporary maintenance and manipulation, retention, retrieval, and prospective use of information, experiences, skills, and intended actions across timescales. \\
Constructed & \emph{Language-Semantic Competence} & 5 & The interpretation and generation of lexical meaning, grammatical and compositional structure, reference, logical-semantic relations, and discourse-level coherence. \\
Constructed & \emph{Language-Pragmatic Competence} & 7 & The interpretation and generation of context-dependent meaning, including implicature, presupposition, indirect speech acts, figurative language, politeness, stance, contextual appropriateness, and sociocultural meaning. \\
Constructed & \emph{Reasoning} & 5 & The derivation and evaluation of deductive, inductive, probabilistic, causal, analogical, and relational inferences. \\
Constructed & \emph{Emotion} & 5 & The observable recognition, interpretation, response to, regulation, acceptance-oriented handling, and use of emotional information concerning oneself, others, or situations. \\
Constructed & \emph{Theory of Mind} & 6 & The representation and inference of agents' emotions, desires, intentions, perceptual perspectives, knowledge states, and beliefs. \\
Integrative & \emph{Metacognition} & 8 & The observable knowledge, monitoring, evaluation, experience, beliefs, and regulation of cognition at individual, collaborative, and human-AI levels. \\
Integrative & \emph{Social Reasoning and Interaction} & 14 & The interpretation and coordination of social information across cues, persons, mental states, communicative intentions, roles, rules, norms, relationships, and collaborative interactions. \\
Integrative & \emph{Planning and Decision-Making} & 14 & An iterative, goal-directed process involving problem framing, self-appraisal, evidence acquisition and evaluation, option generation and selection, action sequencing, implementation, stakeholder and risk management, outcome monitoring, and adaptive revision. \\
Integrative & \emph{Creativity and Innovation} & 7 & The exploratory generation, flexible development, proactive implementation, and iterative refinement of ideas or solutions that are novel, useful, and contextually appropriate. \\
Integrative & \emph{Cultural Competence} & 5 & Critical reflection on cultural assumptions and power, integration of contextual cultural knowledge, adaptive communication, culturally responsive practice, and power-conscious engagement. \\
Integrative & \emph{Moral Reasoning} & 3 & The evaluation of moral situations and the translation of moral judgment into consistent commitment and action, including consideration of principles, norms, consequences, responsibilities, and contextual pressures. \\
\bottomrule
\end{longtable}
\endgroup

\subsection{Complete Domain and Subskill Definitions}

Tables A2-A15 reproduce the full 91-subskill codebook used in the analysis. Definitions specify the conceptual scope of each label; they are not claims about internal model architecture or subjective experience.

\subsubsection{\texorpdfstring{\emph{Perception}}{Perception}}

\textbf{Layer:} Primitive \textbf{Domain definition:} The functional encoding, extraction, integration, organization, recognition, and adaptive interpretation of structure in linguistic, symbolic, sensory, or multimodal input.

\begingroup
\small
\begin{longtable}[]{@{}
>{\raggedright\arraybackslash}p{(\linewidth-2\tabcolsep)*\real{0.280}}
>{\raggedright\arraybackslash}p{(\linewidth-2\tabcolsep)*\real{0.720}}@{}}
\caption{Perception Subskills}\tabularnewline
\toprule
\textbf{Canonical subskill} & \textbf{Definition} \\
\midrule
\endfirsthead
\toprule
\textbf{Canonical subskill} & \textbf{Definition} \\
\midrule
\endhead
Sensory signal transduction & The process of converting physical sensory inputs into neural activity that forms the basis for higher-level perception. \\
Feature extraction \& pattern recognition & Operations that transform early encoded signals into mid-level features and categorical representations and match them to stored templates or statistical models to recognize structure. This sub-skill implements invariant recognition across noise and transformation. \\
Multisensory integration \& cue combination & The integration of sensory information from multiple modalities or cues, weighted by their reliability and prior expectations, to form cohesive percepts. \\
Perceptual organization \& segmentation & Processes that group primitive features into surfaces/objects and segment scenes into meaningful units (figure-ground assignment, closure, continuity), enabling scene structure representation. This is a mid-to-high level perceptual operation distinct from feature extraction (which supplies the features) and from recognition /identification (which labels grouped units). \\
Perceptual learning \& adaptation & Experience-dependent and context-dependent changes in perceptual sensitivity, feature tuning, and cue weighting that persist (learning) or adjust transiently (adaptation) to environmental statistics. This sub-skill supports long-term calibration and short-term normalization of earlier encoding, feature extraction and cue combination processes. \\
\bottomrule
\end{longtable}
\endgroup

\subsubsection{\texorpdfstring{\emph{Attention}}{Attention}}

\textbf{Layer:} Primitive \textbf{Domain definition:} The orienting, selection, and sustained prioritization of task-relevant information while resisting irrelevant, distracting, or competing input.

\begingroup
\small
\begin{longtable}[]{@{}
>{\raggedright\arraybackslash}p{(\linewidth-2\tabcolsep)*\real{0.280}}
>{\raggedright\arraybackslash}p{(\linewidth-2\tabcolsep)*\real{0.720}}@{}}
\caption{Attention Subskills}\tabularnewline
\toprule
\textbf{Canonical subskill} & \textbf{Definition} \\
\midrule
\endfirsthead
\toprule
\textbf{Canonical subskill} & \textbf{Definition} \\
\midrule
\endhead
Sustained attention & The ability to maintain focused attention on tasks or activities for extended periods. \\
Selective attention & The ability to filter relevant stimuli from multiple sources and ignore irrelevant distractions. \\
Orienting attention & The ability to actively orient to and detect stimuli of specific locations or types in the environment. \\
\bottomrule
\end{longtable}
\endgroup

\subsubsection{\texorpdfstring{\emph{Memory}}{Memory}}

\textbf{Layer:} Primitive \textbf{Domain definition:} The encoding, temporary maintenance and manipulation, retention, retrieval, and prospective use of information, experiences, skills, and intended actions across timescales.

\begingroup
\small
\begin{longtable}[]{@{}
>{\raggedright\arraybackslash}p{(\linewidth-2\tabcolsep)*\real{0.280}}
>{\raggedright\arraybackslash}p{(\linewidth-2\tabcolsep)*\real{0.720}}@{}}
\caption{Memory Subskills}\tabularnewline
\toprule
\textbf{Canonical subskill} & \textbf{Definition} \\
\midrule
\endfirsthead
\toprule
\textbf{Canonical subskill} & \textbf{Definition} \\
\midrule
\endhead
Working memory & Working memory refers to the capacity to temporarily hold and manipulate information during ongoing cognitive tasks. \\
Episodic memory & Episodic memory refers to the capacity to recall personal experiences situated in specific temporal and spatial contexts. \\
Procedural memory & Procedural memory refers to the capacity to acquire and perform skills and habits, often without conscious awareness. \\
Prospective memory & Prospective memory refers to the capacity to remember to carry out intended actions in the future. \\
\bottomrule
\end{longtable}
\endgroup

\subsubsection{\texorpdfstring{\emph{Language-Semantic Competence}}{Language-Semantic Competence}}

\textbf{Layer:} Constructed \textbf{Domain definition:} The interpretation and generation of lexical meaning, grammatical and compositional structure, reference, logical-semantic relations, and discourse-level coherence.

\begingroup
\small
\begin{longtable}[]{@{}
>{\raggedright\arraybackslash}p{(\linewidth-2\tabcolsep)*\real{0.280}}
>{\raggedright\arraybackslash}p{(\linewidth-2\tabcolsep)*\real{0.720}}@{}}
\caption{Language-Semantic Competence Subskills}\tabularnewline
\toprule
\textbf{Canonical subskill} & \textbf{Definition} \\
\midrule
\endfirsthead
\toprule
\textbf{Canonical subskill} & \textbf{Definition} \\
\midrule
\endhead
Lexical-semantic knowledge & Understanding word meanings, synonymy, antonymy, hyponymy, polysemy. \\
Compositional semantics & Combining meanings of words and structures into sentence meaning. \\
Referential interpretation & Resolving pronouns, definite descriptions, entities. \\
Logical-semantic reasoning & Understanding entailment, contradiction, quantification, negation. \\
Discourse-level meaning & Integrating meaning across sentences and paragraphs. \\
\bottomrule
\end{longtable}
\endgroup

\subsubsection{\texorpdfstring{\emph{Language-Pragmatic Competence}}{Language-Pragmatic Competence}}

\textbf{Layer:} Constructed \textbf{Domain definition:} The interpretation and generation of context-dependent meaning, including implicature, presupposition, indirect speech acts, figurative language, politeness, stance, contextual appropriateness, and sociocultural meaning.

\begingroup
\small
\begin{longtable}[]{@{}
>{\raggedright\arraybackslash}p{(\linewidth-2\tabcolsep)*\real{0.280}}
>{\raggedright\arraybackslash}p{(\linewidth-2\tabcolsep)*\real{0.720}}@{}}
\caption{Language-Pragmatic Competence Subskills}\tabularnewline
\toprule
\textbf{Canonical subskill} & \textbf{Definition} \\
\midrule
\endfirsthead
\toprule
\textbf{Canonical subskill} & \textbf{Definition} \\
\midrule
\endhead
Implicature inference & Inferring unstated meanings from what is said. \\
Presupposition recognition & Identifying background assumptions triggered by language. \\
Indirect speech act interpretation & Understanding indirect requests, refusals, suggestions. \\
Irony, sarcasm, metaphor, humor & Interpreting non-literal and figurative meanings. \\
Politeness and stance interpretation & Recognizing mitigation, hedging, authority, alignment, disagreement. \\
Contextual appropriateness & Selecting language suitable for audience, genre, relationship, and setting. \\
Sociocultural meaning & Understanding culturally embedded norms, values, and conventions. \\
\bottomrule
\end{longtable}
\endgroup

\subsubsection{\texorpdfstring{\emph{Reasoning}}{Reasoning}}

\textbf{Layer:} Constructed \textbf{Domain definition:} The derivation and evaluation of deductive, inductive, probabilistic, causal, analogical, and relational inferences.

\begingroup
\small
\begin{longtable}[]{@{}
>{\raggedright\arraybackslash}p{(\linewidth-2\tabcolsep)*\real{0.280}}
>{\raggedright\arraybackslash}p{(\linewidth-2\tabcolsep)*\real{0.720}}@{}}
\caption{Reasoning Subskills}\tabularnewline
\toprule
\textbf{Canonical subskill} & \textbf{Definition} \\
\midrule
\endfirsthead
\toprule
\textbf{Canonical subskill} & \textbf{Definition} \\
\midrule
\endhead
Deductive reasoning & The application of formal logical rules to derive conclusions that follow necessarily from premises. \\
Inductive reasoning & The process of forming generalizations and hypotheses from specific instances or observations, generating conclusions that may be probabilistic or tentative. \\
Probabilistic/Bayesian reasoning & The process of updating beliefs or predictions in the face of uncertainty using probabilistic principles, such as Bayes' theorem, to revise and refine judgments. \\
Causal reasoning (intervention \& counterfactuals) & The process of inferring causal relationships and testing hypotheses through counterfactual thinking, interventions, and exploring cause-effect scenarios. \\
Analogical and relational reasoning & The capacity to map relational structure from a source to a target (analogy) and to reason about relations among elements (relational complexity), supporting transfer, explanation, and problem-solving. \\
\bottomrule
\end{longtable}
\endgroup

\subsubsection{\texorpdfstring{\emph{Emotion}}{Emotion}}

\textbf{Layer:} Constructed \textbf{Domain definition:} The observable recognition, interpretation, response to, regulation, acceptance-oriented handling, and use of emotional information concerning oneself, others, or situations.

\begingroup
\small
\begin{longtable}[]{@{}
>{\raggedright\arraybackslash}p{(\linewidth-2\tabcolsep)*\real{0.280}}
>{\raggedright\arraybackslash}p{(\linewidth-2\tabcolsep)*\real{0.720}}@{}}
\caption{Emotion Subskills}\tabularnewline
\toprule
\textbf{Canonical subskill} & \textbf{Definition} \\
\midrule
\endfirsthead
\toprule
\textbf{Canonical subskill} & \textbf{Definition} \\
\midrule
\endhead
Emotion regulation & The ability to flexibly adjust the intensity, duration, and expression of emotions. \\
Emotion perception and recognition & The ability to detect one's own and others' emotions, including identifying emotional states through nonverbal signals such as facial expressions and tone of voice. \\
Emotion Understanding and Clarity & The ability to comprehend the causes of emotions, distinguish between different types of emotions, and clearly grasp the nature of one's own emotional experiences. \\
Emotion Acceptance & The ability to face one's own emotional experiences (especially negative emotions) and avoid secondary distress caused by emotional reactions. \\
Emotion Utilization & The ability to use emotions to promote thinking, creativity, motivation, and problem-solving. \\
\bottomrule
\end{longtable}
\endgroup

\subsubsection{\texorpdfstring{\emph{Theory of Mind}}{Theory of Mind}}

\textbf{Layer:} Constructed \textbf{Domain definition:} The representation and inference of agents' emotions, desires, intentions, perceptual perspectives, knowledge states, and beliefs.

\begingroup
\small
\begin{longtable}[]{@{}
>{\raggedright\arraybackslash}p{(\linewidth-2\tabcolsep)*\real{0.280}}
>{\raggedright\arraybackslash}p{(\linewidth-2\tabcolsep)*\real{0.720}}@{}}
\caption{Theory of Mind Subskills}\tabularnewline
\toprule
\textbf{Canonical subskill} & \textbf{Definition} \\
\midrule
\endfirsthead
\toprule
\textbf{Canonical subskill} & \textbf{Definition} \\
\midrule
\endhead
ToM of emotion & ToM of Emotion refers to the ability to recognize and interpret emotional states in oneself and others based on facial expressions, body language, and contextual cues. It involves understanding how emotions can influence behavior and interpersonal interactions, allowing individuals to respond appropriately to the emotional needs of others. It is essential for fostering empathy and social cohesion. \\
ToM of desirability & ToM of desirability involves the capacity to discern and interpret the desires or preferences of oneself and others. It enables individuals to understand what motivates others' actions based on their wants and needs, which may not always be explicitly expressed. It is crucial for predicting future behaviors and facilitating cooperative interactions. \\
ToM of intention & ToM of Intention pertains to the ability to recognize and infer the intentions behind others' actions. It enables individuals to understand that actions are often goal-directed and based on underlying motivations. It allows for the anticipation of behaviors and is key in enabling effective communication and collaboration. \\
Perspective Taking and Percepts & Perspective Taking is the cognitive skill that allows individuals to understand and appreciate another person's viewpoint or mental state. This involves recognizing that others may have different thoughts, feelings, and experiences than oneself. It is foundational for empathy, negotiation, and conflict resolution, as it encourages understanding and acceptance of diverse perspectives. \\
ToM of Knowledge & ToM of Knowledge refers to the ability to understand what others know or believe about a given situation, especially when that knowledge is incomplete or differs from one's own. This sub-skill is crucial for effective communication, as it guides individuals in sharing information appropriately and recognizing when misunderstandings may occur. \\
ToM of Beliefs & ToM of Beliefs involves the capacity to comprehend that others can hold beliefs that may be false or different from reality. This sub-skill is essential for understanding scenarios involving deception, misinformation, and divergent viewpoints, allowing individuals to navigate complex social interactions with a nuanced understanding of others' mental states. \\
\bottomrule
\end{longtable}
\endgroup

\subsubsection{\texorpdfstring{\emph{Metacognition}}{Metacognition}}

\textbf{Layer:} Integrative \textbf{Domain definition:} The observable knowledge, monitoring, evaluation, experience, beliefs, and regulation of cognition at individual, collaborative, and human-AI levels.

\begingroup
\small
\begin{longtable}[]{@{}
>{\raggedright\arraybackslash}p{(\linewidth-2\tabcolsep)*\real{0.280}}
>{\raggedright\arraybackslash}p{(\linewidth-2\tabcolsep)*\real{0.720}}@{}}
\caption{Metacognition Subskills}\tabularnewline
\toprule
\textbf{Canonical subskill} & \textbf{Definition} \\
\midrule
\endfirsthead
\toprule
\textbf{Canonical subskill} & \textbf{Definition} \\
\midrule
\endhead
Critical Thinking / Evaluative Reasoning & Analyzing and evaluating solution options to make recommendations/decisions. \\
Metacognitive knowledge & Awareness and stored information about one's own cognitive abilities, strategies, and tasks. \\
Metacognitive monitoring & Real-time self-observation and assessment of cognitive states and performance. \\
Metacognitive regulation & The deliberate management of cognition through the selection, initiation, and adaptation of strategies. \\
Metacognitive experiences & Subjective feelings and judgments that accompany cognition. \\
Metacognitive Attitudes and Beliefs & Convictions about the nature of knowledge and learning that shape cognitive engagement. \\
Social and Collaborative Metacognition & The processes by which individuals monitor, communicate, and co-regulate thinking in group settings. \\
Human-AI Metacognition & The capacity to plan, monitor, and reflect on one's own cognition while collaborating with an AI partner. \\
\bottomrule
\end{longtable}
\endgroup

\subsubsection{\texorpdfstring{\emph{Social Reasoning and Interaction}}{Social Reasoning and Interaction}}

\textbf{Layer:} Integrative \textbf{Domain definition:} The interpretation and coordination of social information across cues, persons, mental states, communicative intentions, roles, rules, norms, relationships, and collaborative interactions.

\begingroup
\small
\begin{longtable}[]{@{}
>{\raggedright\arraybackslash}p{(\linewidth-2\tabcolsep)*\real{0.280}}
>{\raggedright\arraybackslash}p{(\linewidth-2\tabcolsep)*\real{0.720}}@{}}
\caption{Social Reasoning and Interaction Subskills}\tabularnewline
\toprule
\textbf{Canonical subskill} & \textbf{Definition} \\
\midrule
\endfirsthead
\toprule
\textbf{Canonical subskill} & \textbf{Definition} \\
\midrule
\endhead
Knowledge sharing \& communication & Exchanging information/skills with others as part of carrying out creative work. \\
Social appropriateness judgment & Ability to classify a behavior in context as appropriate vs.~inappropriate relative to expected social conduct. \\
Social-rule representation and domain differentiation & Knowledge of social rules and the ability to distinguish moral rules (welfare/rights; context-free) from conventional rules (authority-/context-dependent), which structures normative social reasoning. \\
Ecological norm comprehension (interpersonal \& intrapersonal) & Understanding whether a character's behavior matches how people should behave (interpersonal norm comprehension) and whether oneself would behave similarly (intrapersonal norm comprehension). \\
Social cue identification & Ability to locate and report the relevant cue(s) in a scene that supports the judgment. \\
Teamwork (collaborative co-creation) & Working interdependently to generate more responses and solutions. \\
Contextual integration of social information & Ability to extract relevant social information and integrate it with contextual factors to form an optimal interpretation of the interaction dynamics (beyond cue listing). \\
Recursive / higher-order mental-state reasoning & Reasoning about embedded mental states (e.g., second-order beliefs) that support more advanced social reasoning in interactions, especially when one person's mental state must be represented within another's. \\
Communicative-intention inference in pragmatic contexts & Inferring a speaker's intended meaning when it diverges from literal meaning (e.g., irony), including detecting the belief/intention structure that supports that inference. \\
Other-perspective inference & Ability to represent someone else's perspective as distinct from one's own. \\
Self-perspective inhibition (egocentricity control) & Executive ability to suppress one's own perspective/knowledge when it conflicts with (and would contaminate) reasoning about another person's mental state; separable from perspective inference itself. \\
Access and use of social knowledge schemas to facilitate inference & Using stored social knowledge/pragmatic schemas to improve reasoning performance in social contexts (a facilitation that can break down with frontal-lobe dysfunction). \\
Self-referential projection as a strategy & Tendency to use the self as a reference point when evaluating real social situations-often a prominent strategy in social reasoning, though it can bias inference. \\
Social and person perception & Social and person perception refers to the ability to extract and infer socially relevant information (faces, emotion, identity, intentions) from perceptual inputs, integrating sensory cues with social-cognitive priors; it operates as a specialized domain that recruits both perceptual and social-cognitive systems and emphasizes social meaning and rapid influence of social priors. \\
\bottomrule
\end{longtable}
\endgroup

\subsubsection{\texorpdfstring{\emph{Planning and Decision-Making}}{Planning and Decision-Making}}

\textbf{Layer:} Integrative \textbf{Domain definition:} An iterative, goal-directed process involving problem framing, self-appraisal, evidence acquisition and evaluation, option generation and selection, action sequencing, implementation, stakeholder and risk management, outcome monitoring, and adaptive revision.

\begingroup
\small
\begin{longtable}[]{@{}
>{\raggedright\arraybackslash}p{(\linewidth-2\tabcolsep)*\real{0.280}}
>{\raggedright\arraybackslash}p{(\linewidth-2\tabcolsep)*\real{0.720}}@{}}
\caption{Planning and Decision-Making Subskills}\tabularnewline
\toprule
\textbf{Canonical subskill} & \textbf{Definition} \\
\midrule
\endfirsthead
\toprule
\textbf{Canonical subskill} & \textbf{Definition} \\
\midrule
\endhead
Problem framing and goal specification & Ability to define the decision problem clearly, articulate objectives, and translate them into actionable decision questions or goals (e.g., "define a practice question" or "goal selection"). \\
Self-appraisal and capability-demand alignment & Ability to evaluate one's own competencies, constraints, and values and judge fit with the task context; supports realistic goal setting and confident commitment to a choice. \\
Information acquisition and evidence search & Ability to identify what information is needed and systematically search for relevant evidence (research, guidelines, contextual data) to reduce uncertainty before choosing. \\
Evidence appraisal and credibility judgment & Ability to critically appraise evidence quality (strength, applicability) and differentiate robust evidence from weak or non-generalizable sources. \\
Synthesis and recommendation formation & Ability to integrate multiple evidence sources into coherent interpretations and derive defensible recommendations rather than isolated facts. \\
Contextual adaptation and constraint handling & Ability to adapt evidence-based recommendations to real-world constraints (local context, resources, political/community preferences) so that chosen plans are feasible. \\
Option generation and alternative construction & Ability to develop viable courses of action (not just pick from given choices), ensuring alternatives are meaningfully distinct and implementable within constraints. \\
Evaluation, comparison, and selection (decision rule use) & Ability to compare alternatives against explicit criteria (effectiveness, fit, risk/effort) and commit to a selected option with a justified rationale. \\
Action planning and sequencing & Ability to convert the chosen option into an action plan (steps, sequencing, responsibilities, timing), moving from "decision" to "execution design.". \\
Barrier/facilitator diagnosis and risk management & Ability to anticipate barriers and enabling conditions (resources, organizational factors, guideline/evidence constraints, stakeholder values) and plan mitigations proactively. \\
Stakeholder analysis and engagement planning & Ability to map stakeholders' influence/importance, coordinate buy-in, and design engagement steps that support adoption and reduce implementation friction. \\
Implementation execution (plan enactment) & Ability to implement adapted evidence/decisions in practice, including mobilizing people and resources to carry out the plan as intended. \\
Outcome evaluation and indicator-based monitoring & Ability to define evaluation indicators and assess whether the implemented decision produced intended outcomes; supports iterative improvement and accountability. \\
Problem solving and adaptive revision & Ability to respond to obstacles during execution, revise plans when assumptions fail, and maintain progress toward goals under changing conditions. \\
\bottomrule
\end{longtable}
\endgroup

\subsubsection{\texorpdfstring{\emph{Creativity and Innovation}}{Creativity and Innovation}}

\textbf{Layer:} Integrative \textbf{Domain definition:} The exploratory generation, flexible development, proactive implementation, and iterative refinement of ideas or solutions that are novel, useful, and contextually appropriate.

\begingroup
\small
\begin{longtable}[]{@{}
>{\raggedright\arraybackslash}p{(\linewidth-2\tabcolsep)*\real{0.280}}
>{\raggedright\arraybackslash}p{(\linewidth-2\tabcolsep)*\real{0.720}}@{}}
\caption{Creativity and Innovation Subskills}\tabularnewline
\toprule
\textbf{Canonical subskill} & \textbf{Definition} \\
\midrule
\endfirsthead
\toprule
\textbf{Canonical subskill} & \textbf{Definition} \\
\midrule
\endhead
Intrinsic creative motivation & An internally driven desire to pursue novel-and-useful solutions. \\
Curiosity / intellectual openness & An active tendency to explore, question, and seek information/alternatives. \\
Creative ideation capacity & Generating new ideas or methods to solve complex problems. \\
Cognitive flexibility (creative thinking style) & Shifting perspectives/categories and integrating information from diverse sources. \\
Initiative / proactive enactment & Taking actions to implement new ideas rather than stopping at ideation. \\
Boundary-pushing problem solving & Observing/diagnosing problems and crossing existing boundaries to craft workable solutions. \\
Innovative behavior cycle & The ability to generate, develop, apply, promote/carry out, and modify new ideas. \\
\bottomrule
\end{longtable}
\endgroup

\subsubsection{\texorpdfstring{\emph{Cultural Competence}}{Cultural Competence}}

\textbf{Layer:} Integrative \textbf{Domain definition:} Critical reflection on cultural assumptions and power, integration of contextual cultural knowledge, adaptive communication, culturally responsive practice, and power-conscious engagement.

\begingroup
\small
\begin{longtable}[]{@{}
>{\raggedright\arraybackslash}p{(\linewidth-2\tabcolsep)*\real{0.280}}
>{\raggedright\arraybackslash}p{(\linewidth-2\tabcolsep)*\real{0.720}}@{}}
\caption{Cultural Competence Subskills}\tabularnewline
\toprule
\textbf{Canonical subskill} & \textbf{Definition} \\
\midrule
\endfirsthead
\toprule
\textbf{Canonical subskill} & \textbf{Definition} \\
\midrule
\endhead
Critical cultural self-reflexivity & The ongoing cognitive process of identifying, interrogating, and mitigating one's own cultural assumptions, implicit biases, social privileges, and positional power that may unconsciously shape professional perceptions, decisions, or interactions with individuals from different cultural backgrounds. \\
Contextual cultural knowledge integration & The ability to accurately acquire, critically evaluate, and ethically apply dynamic, non-stereotypical knowledge about the historical, social, spiritual, linguistic, and health-related norms of specific cultural groups-while recognizing heterogeneity within groups and avoiding essentialism. \\
Adaptive Communication & The real-time behavioral skill of adjusting verbal expression, nonverbal cues (e.g., eye contact, personal space, silence), language complexity, and use of interpretation resources to match the culturally informed communication preferences of the service user, thereby enhancing clarity, trust, and mutual understanding. \\
Culturally responsive practice adaptation & The procedural competence to intentionally modify assessment protocols, intervention plans, scheduling, environmental arrangements, or goal-setting processes to align with the client's cultural values, priorities, and lived context-without diluting clinical or professional integrity. \\
Power-conscious engagement & The relational skill of recognizing asymmetries in authority, voice, and access within cross-cultural professional encounters-and actively redistributing interactional power through collaborative decision-making, validating client expertise, and challenging institutional barriers to equity. \\
\bottomrule
\end{longtable}
\endgroup

\subsubsection{\texorpdfstring{\emph{Moral Reasoning}}{Moral Reasoning}}

\textbf{Layer:} Integrative \textbf{Domain definition:} The evaluation of moral situations and the translation of moral judgment into consistent commitment and action, including consideration of principles, norms, consequences, responsibilities, and contextual pressures.

\begingroup
\small
\begin{longtable}[]{@{}
>{\raggedright\arraybackslash}p{(\linewidth-2\tabcolsep)*\real{0.280}}
>{\raggedright\arraybackslash}p{(\linewidth-2\tabcolsep)*\real{0.720}}@{}}
\caption{Moral Reasoning Subskills}\tabularnewline
\toprule
\textbf{Canonical subskill} & \textbf{Definition} \\
\midrule
\endfirsthead
\toprule
\textbf{Canonical subskill} & \textbf{Definition} \\
\midrule
\endhead
Moral Implementation (Courage and Action) & The behavioral execution of a moral decision in the face of fear, risk, or opposition-requiring perseverance, ego strength, and resilience to translate judgment into action. \\
Moral Judgment (Cognitive Evaluation) & The model's ability to evaluate moral scenarios by classifying actions, intentions, or outcomes along ethical dimensions. This includes detecting harm and intent, applying moral principles (care, fairness, authority, etc.), and reasoning through ethical trade-offs. It measures how well the model maps a situation to an appropriate moral label, reflecting both its ethical knowledge and its capacity for structured reasoning about right and wrong. \\
Moral Motivation (Integrity / Commitment) & The model's consistency between its stated moral principles and its actual outputs across contexts. This evaluates whether the model upholds its ethical commitments when faced with adversarial prompts, role-playing scenarios, or pressure to produce harmful content. It captures the model's resistance to moral hypocrisy-saying the right thing but failing to act accordingly when probed. Core test: does the model stand by its moral judgments under stress or in ambiguous situations? \\
\bottomrule
\end{longtable}
\endgroup

\end{document}